\documentclass{neus2025}

\usepackage{times}
\usepackage{booktabs}
\usepackage{tcolorbox}
\tcbuselibrary{breakable}
\usepackage{multicol}
\usepackage{enumitem}
\usepackage{wrapfig}
\usepackage{colortbl}
\usepackage{multirow}
\usepackage{listings}
\usepackage{array}
\usepackage{longtable}
\usepackage[bottom]{footmisc}   

\usepackage[format=plain, justification=justified, singlelinecheck=true, margin=0pt]{caption}
\usepackage{tikz}
\usetikzlibrary{shapes,arrows,positioning,arrows.meta,shadows.blur,calc,decorations.pathmorphing,patterns}

\definecolor{teal700}{HTML}{0F766E}
\definecolor{teal50}{HTML}{F0FDFA}
\definecolor{slate700}{HTML}{334155}
\definecolor{slate900}{HTML}{0F172A}
\definecolor{slate500}{HTML}{64748B}
\definecolor{slate200}{HTML}{E2E8F0}
\definecolor{slate50}{HTML}{F8FAFC}
\definecolor{amber600}{HTML}{D97706}
\definecolor{amber50}{HTML}{FFFBEB}
\definecolor{rose700}{HTML}{BE123C}

\newtcolorbox{insightbox}[1][]{colback=blue!5, colframe=blue!50!black, title=\textbf{Key Insight}, breakable, #1}
\newtcolorbox{scenariobox}[1][]{colback=green!5, colframe=green!50!black, title=\textbf{Scenario}, breakable, #1}
\newtcolorbox{exercisebox}[1][]{colback=yellow!10, colframe=orange!80!black, title=\textbf{Exercise}, breakable, #1}

\newcommand{\Lcontra}{\mathcal{L}_{\text{contra}}}

\newcommand{\smin}{\mathrm{smin}_\tau}
\newcommand{\smax}{\mathrm{smax}_\tau}
\newcommand{\relu}{\mathrm{ReLU}}
\newcommand{\tm}{\textsf{torchmodal}}

\newenvironment{glance}[1]{%
  \par\smallskip\small
  \setlength{\LTleft}{0pt}\setlength{\LTpre}{0pt}\setlength{\LTpost}{0pt}%
  \begin{longtable}{@{}>{\raggedright\arraybackslash\bfseries}p{0.17\linewidth}>{\raggedright\arraybackslash}p{0.79\linewidth}@{}}
  \toprule
  \multicolumn{2}{@{}l}{\textbf{#1}}\\
  \midrule\endfirsthead
  \toprule
  \multicolumn{2}{@{}l}{\textbf{#1} (continued)}\\
  \midrule\endhead
  \bottomrule\endfoot
  \bottomrule\endlastfoot}{%
  \end{longtable}\addtocounter{table}{-1}\par\smallskip}

\title[Differentiable Modal Logic for Multi-Agent Systems]{Differentiable Modal Logic for Multi-Agent Diagnosis, Orchestration and Communication}

\author{%
  \Name{Antonin Sulc}
  \Email{asulc@lbl.gov}\\
  \addr Lawrence Berkeley National Lab\\
  \addr Berkeley, U.S.A.
}

\begin{document}

\maketitle

\raggedbottom
\emergencystretch=3em

\begin{abstract}
As multi-agent AI systems evolve from simple chatbots to autonomous swarms, debugging semantic failures requires reasoning about knowledge, belief, causality, and obligation, precisely what modal logic was designed to formalize. However, traditional modal logic requires manual specification of relationship structures that are unknown or dynamic in real systems. This tutorial demonstrates differentiable modal logic (DML), implemented via Modal Logic Neural Networks (MLNNs), enabling systems to learn trust networks, temporal orders, and regulatory boundaries from behavioral data.

We present a unified neurosymbolic debugging framework through four modalities: epistemic (who to trust), temporal (what can follow what), deontic (what actions are permitted), and doxastic (who believes what). Each modality is demonstrated on concrete multi-agent scenarios (from exposing an attacker in a drone swarm to deciding when to consult another LLM code reviewer), with complete implementations showing how logical contradictions become learnable optimization objectives. Key contributions for the neurosymbolic community: (1) interpretable learned structures where trust and temporal order are explicit parameters, not opaque embeddings; (2) knowledge injection via differentiable axioms that guide learning with sparse data; (3) compositional multi-modal reasoning that combines epistemic and doxastic constraints; and (4) practical deployment patterns for monitoring, active control and communication of multi-agent systems. All code is provided as executable Python scripts.
\end{abstract}


\begin{figure}[t]
\centering
\resizebox{\textwidth}{!}{%
\begin{tikzpicture}[
    font=\sffamily,
    >=Stealth,
    world/.style={circle, draw=teal700, line width=0.9pt, fill=teal50,
        minimum size=1.55cm, text=teal700, font=\LARGE},
    embed/.style={circle, draw=teal700, line width=0.7pt, fill=white,
        minimum size=0.85cm, text=teal700, font=\normalsize},
    val_box/.style={rectangle, draw=teal700, line width=0.6pt, fill=teal50!50,
        rounded corners=1.0mm, minimum width=2.4cm, minimum height=0.78cm,
        font=\large, align=center},
    op_box/.style={rectangle, draw=slate700, fill=slate50,
        rounded corners=1.5mm, minimum width=6.2cm, minimum height=1.4cm,
        font=\Large, align=center},
    bound_box/.style={rectangle, draw=amber600, line width=0.7pt, fill=amber50,
        rounded corners=1.0mm, minimum width=3.7cm, minimum height=0.85cm,
        font=\large, align=center},
    learn_strong/.style={->, line width=1.7pt, draw=rose700,
        shorten >=2pt, shorten <=2pt},
    learn_mid/.style={->, line width=1.05pt, draw=rose700!75,
        shorten >=2pt, shorten <=2pt},
    learn_weak/.style={->, dashed, semithick, draw=rose700!45,
        shorten >=2pt, shorten <=2pt},
    flow/.style={->, semithick, draw=slate500, rounded corners=1.4mm},
    grad/.style={->, line width=1.8pt, dashed, draw=amber600},
    panel_label/.style={font=\Large\bfseries, text=slate700},
    encode/.style={->, semithick, dashed, draw=slate700!55,
        shorten >=2pt, shorten <=2pt},
    axiom/.style={rectangle, rounded corners=1.0mm, draw=slate700,
        fill=slate50, line width=0.6pt, inner sep=4pt,
        font=\LARGE\bfseries, text=slate900},
]
\colorlet{slate500}{slate700!75}

\node[panel_label] at (3.0, 4.7) {Reasoning over worlds};
\node[world] (w1) at (1.4, 2.7) {$a$};
\node[world] (w2) at (4.6, 2.7) {$b$};
\node[world] (w3) at (3.0, 0.9) {$c$};
\node[val_box, anchor=south, yshift=2mm] (vw1) at (w1.north) {waits at red};
\node[val_box, anchor=south, yshift=2mm] (vw2) at (w2.north) {drives on green};
\node[val_box, anchor=north, yshift=-2mm] (vw3) at (w3.south) {drives through red};
\draw[learn_strong] (w1) -- node[above=0.5mm, font=\large, text=rose700] {$A_\theta{=}0.9$} (w2);
\draw[learn_mid] (w1) -- node[left=1mm, font=\large, text=rose700!85] {$A_\theta{=}0.6$} (w3);
\draw[learn_strong] (w1) edge[in=200, out=160, looseness=14]
    node[left=0.5mm, font=\large, text=rose700] {$A_\theta{=}1.0$} (w1);
\node[panel_label] at (10.5, 4.7) {The rule at $a$ (Eq.~\ref{eq:necessity})};
\node[op_box, fill=amber50, draw=amber600, minimum height=2.0cm, minimum width=7.0cm, font=\Large]
    (loss) at (10.5, 0.9)
    {$\Box(\mathit{red}\to\neg\mathit{move})(a) \approx 0.40$\\[3pt]
     {\small $c$ breaks the rule and is permitted at $0.6$:\; $\mathcal{L}_{\text{contra}} = 0.60$ lowers $A_\theta[a,c]$}};
\draw[flow] (w2.east) to[bend left=18] (loss.north west);
\draw[flow] (w3.east) to[bend right=8] (loss.south west);

\begin{scope}[xshift=17cm]
\node[panel_label] at (3.0, 4.7) {Worlds as embeddings};
\fill[slate50]
    plot[smooth cycle, tension=0.7] coordinates
    {(-0.05,0.4) (1.4,-0.3) (3.5,-0.35) (5.4,0.25) (6.3,1.3)
     (6.2,2.7) (5.0,3.5) (3.0,3.7) (1.0,3.4) (0.0,2.2) (-0.25,1.0)};
\fill[rose700!11]
    plot[smooth cycle, tension=0.65] coordinates
    {(0.35,0.55) (1.6,0.0) (3.5,-0.05) (5.2,0.5) (5.95,1.5)
     (5.75,2.85) (4.5,3.35) (2.6,3.45) (0.85,3.1) (0.15,2.0) (-0.05,1.1)};
\fill[teal700!18]
    plot[smooth cycle, tension=0.7] coordinates
    {(0.85,1.0) (1.85,0.5) (3.5,0.55) (4.8,1.0) (5.2,1.85)
     (4.95,2.85) (3.65,3.15) (2.0,3.05) (0.95,2.55) (0.55,1.65)};
\fill[teal700!32]
    plot[smooth cycle, tension=0.7] coordinates
    {(1.4,1.4) (2.3,1.0) (3.6,1.05) (4.35,1.5) (4.55,2.45)
     (3.5,2.9) (2.0,2.85) (1.3,2.2)};
\fill[teal700!55]
    plot[smooth cycle, tension=0.7] coordinates
    {(1.95,1.85) (2.65,1.65) (3.4,1.85) (3.4,2.4) (2.65,2.65) (1.9,2.45)};
\node[font=\large\bfseries, text=white] at (3.05, 2.55) {$\Box\phi$};
\node[font=\normalsize\itshape, text=teal700!90] at (4.55, 1.55) {$\phi$ holds};
\node[font=\normalsize\itshape, text=rose700!85] at (5.30, 2.85) {$\Diamond\phi$ holds};
\node[embed] (h1) at (2.45, 2.05) {$h_a$};
\node[embed] (h2) at (4.05, 2.20) {$h_b$};
\node[embed] (h3) at (0.55, 0.95) {$h_c$};
\node[font=\normalsize\itshape, text=slate700, align=center] at (3.0, -0.7)
    {with a reflexive relation:\\
     $\Box\phi$ worlds $\subseteq$ $\phi$ worlds $\subseteq$ $\Diamond\phi$ worlds};
\end{scope}

\draw[encode] (w1.north east) to[bend left=8]
    node[midway, above=2pt, font=\large\itshape, text=slate700] {$f_\theta$}
    (h1.west);
\end{tikzpicture}}
\caption{\textbf{Overview, on the running example of Sect.~\ref{sec:foundations}.} Left: a car waits at a red light ($a$); the worlds are the situations it can be in next, and the learned accessibility $A_\theta$ says how far each next situation is permitted. Middle: the rule of the road $\Box(\mathit{red}\to\neg\mathit{move})$ at $a$ (Eq.~\ref{eq:necessity}), a smooth minimum over the permitted next situations. Driving through on red ($c$) breaks the rule and is still permitted at $0.6$, so the rule holds only to $0.40$; its contradiction loss (Sect.~\ref{subsec:loss}) is $0.60$, and its gradient lowers $A_\theta[a,c]$. Right: the metric form of the relation (Sect.~\ref{subsec:accessibility}), in which each situation is embedded as $h_w = f_\theta(x_w)$ and $A_\theta$ is computed from the embeddings, so a new situation gets a row as soon as its features are known. The shaded regions illustrate, after training, the worlds where $\Box\phi$, $\phi$ and $\Diamond\phi$ hold, for $\phi = \mathit{red}\to\neg\mathit{move}$; $c$ lies outside $\phi$. The regions are nested when the relation is reflexive (axiom T, Table~\ref{tab:frame_axioms}).}
\label{fig:teaser}
\end{figure}

\vspace{-1em}\section{Introduction}\vspace{-0.25em}
\label{sec:intro}

An execution log records what agents did, but leaves other questions open. A message from agent A to agent B does not establish whether B understood it. A completed action does not establish whether it was permitted, and a recorded approval does not establish whether the reviewer had the authority to give it. Answering these questions requires a model of what agents know or believe, which actions are permitted, and how events unfold over time. Epistemic, doxastic, deontic, and temporal logics provide ways to express these distinctions (see~Table~\ref{tab:modalities}).

These logics evaluate a formula over possible worlds linked by an accessibility relation (Kripke semantics). The relation is usually the hard part to write down. For a robot, it depends on which situations its camera can tell apart; for permitted actions, on the rules of the task; in code review, on whose judgement should count.

Modal Logic Neural Networks (MLNNs)~\citep{sulc2026modal} learn the relation instead. It becomes a matrix $A_\theta$ with entries in $[0,1]$, $\Box$ and $\Diamond$ become smooth minima and maxima over it, and each axiom becomes a loss that is zero when the axiom holds. Training lowers these losses; no entry of the relation is labelled.

\vspace{-1em}\subsection{Outline}\vspace{-0.25em}

Sects.~\ref{sec:foundations} and~\ref{sec:method} introduce Kripke models and the MLNN of \citet{sulc2026modal} on a traffic-light example, which readers can work through by hand and run in \textsf{torchmodal}.\footnote{\textsf{torchmodal} is on pip (\texttt{pip install torchmodal}) and at \url{https://github.com/sulcantonin/torchmodal}.}

Sects.~\ref{sec:knowledge}--\ref{sec:tom} then apply the same training loop to four problems, changing the interpretation of the modal operators in each case (Table~\ref{tab:modalities}). In the warehouse example, robots' reports provide the evidence for learning what they know, including the distinction between knowledge and a faulty robot's beliefs (Sect.~\ref{sec:knowledge}). The temporal example learns which sentences can follow others in a story (Sect.~\ref{sec:temporal}). The deontic example infers permitted car movements from examples of compliant driving and one written rule (Sect.~\ref{sec:deontic}). Finally, a network learns from a story who witnessed each event; nested $\Box$ operators then answer questions such as ``where does Anne think Bob thinks the ball is?'' (Sect.~\ref{sec:tom}).

The final two examples combine modal operators to guide agents' decisions (Sects.~\ref{sec:orchestration} and~\ref{sec:communication}). A code-review pipeline merges a change when $\Box\,\mathit{correct}$ holds, rejects it when $\Box\,\neg\mathit{correct}$ holds, and otherwise consults a more expensive LLM reviewer. Finally, a team of drones learns who would see what, checks each message against what it sees, and from the failed checks learns how far to trust each drone; with that trust it maps the terrain and runs a relay around the obstacles (Sect.~\ref{sec:communication}).

The case studies are illustrations, not benchmarks. On most tasks an established, simpler method does as well, and sometimes better; on the last, a simple ranking comes close (Table~\ref{tab:studies}). The modal model's advantage is interpretability: the relation is learned from data (induction), can be inspected entry by entry and checked against the frame axioms (Sect.~\ref{subsec:audit}), and answers new formulas by deduction, including questions it was never trained on. Sect.~\ref{sec:limitations} discusses the limitations. The core of each study is listed in Sect.~\ref{app:code}, and the accompanying code reproduces every number reported in the paper.
\vspace{-1em}\section{Related Work}\vspace{-0.25em}
\label{sec:related}

\paragraph{Neurosymbolic learning.}
Several neurosymbolic approaches make logical reasoning part of neural-network training~\citep{garcez2023neurosymbolic, colelough2025neurosymbolic}. They differ in how they represent formulas and use them during learning. Logic Tensor Networks~\citep{badreddine2022logic} evaluate first-order formulas under fuzzy semantics. DeepProbLog~\citep{manhaeve2018deepproblog} and Scallop~\citep{li2023scallop} support differentiable probabilistic logic programming, while semantic loss~\citep{xu2018semantic} turns propositional constraints into a training objective. Logical Neural Networks (LNNs)~\citep{riegel2020logical} propagate lower and upper truth bounds through weighted logical neurons.

MLNNs~\citep{sulc2026modal} extend the bound-based construction of LNNs to modal operators and allow the accessibility relation to be learned as an explicit matrix. Earlier connectionist modal logic~\citep{garcez2007connectionist} translates modal logic programs into neural-network ensembles, with the accessibility relation given. The soft operators used here come from many-valued modal logic, which treats graded accessibility~\citep{fitting1991many, hajek1998metamathematics}; \citet{vankrieken2022analyzing} analyse how such differentiable fuzzy operators behave.

\paragraph{Modal logic in multi-agent systems.}
Epistemic and doxastic logics distinguish knowledge from belief: knowledge entails truth, whereas belief can be mistaken~\citep{hintikka1962knowledge, fagin2003reasoning}. Modal accounts also connect belief to information acquisition and trust~\citep{liau2003belief}. Tools such as mlsolver~\citep{mlsolver} evaluate modal formulas on a Kripke structure that the user supplies. Our examples learn that structure from data, and the formulas are then evaluated on what was learned.

Trust and reputation models are another way to weigh information from other agents~\citep{ramchurn2004trust}; the beta reputation system~\citep{josang2002beta}, for example, aggregates feedback through beta distributions. In our examples the weight of an agent's reports is read from a learned relation instead: each reviewer of Sect.~\ref{sec:orchestration} has its own, and the drones of Sect.~\ref{sec:communication} learn a trust value per drone from the contradictions between their messages.

\paragraph{Temporal properties and norms.}
Model checking~\citep{baier2008principles} verifies temporal properties against a specified transition system, while runtime verification~\citep{bauer2011rv} checks properties against observations collected during execution, and STLCG~\citep{leung2023backpropagation} back-propagates through the quantitative semantics of signal temporal logic. None of them learns which transitions exist. Sects.~\ref{sec:temporal} and~\ref{sec:deontic} learn it instead: which transitions are plausible (temporal) and which are permitted (deontic). The two differ, because an observed transition need not be a permitted one.

\paragraph{Theory of mind.}
Story-based theory-of-mind benchmarks ask readers to track what characters believe as people enter or leave rooms and objects change location~\citep{le2019revisiting}. Hi-ToM~\citep{wu2023hitom} includes beliefs nested up to fourth order and reports declining performance among the evaluated language models as the reasoning depth increases. Other work tracks beliefs explicitly: SymbolicToM~\citep{sclar2023minding} maintains representations of characters' beliefs, including beliefs about other characters.

Sect.~\ref{sec:tom} instead learns who observed which events, as the relation, from answers to first-order questions only. Nested modal operators like $\Box_{\text{Anne}}\Box_{\text{Bob}}\,\mathit{loc}_c$ (``Anne thinks Bob thinks the ball is in $c$'') then answer the higher-order questions from the same relation.

\paragraph{Combining LLM judgements.}
LLM-based evaluation uses model judgements to assess other models' outputs~\citep{zheng2023judging, gu2024survey}. The statistical problem of combining fallible judgements predates LLMs. \citet{dawid1979maximum} estimate annotator-specific error rates using confusion matrices. Sect.~\ref{sec:orchestration} considers a related question within a modal formulation: which reviewer judgements should count towards accepting or rejecting a change, and when should another reviewer be consulted?
\section{Kripke Models}
\label{sec:background}\label{sec:foundations}

\subsection{Worlds, accessibility, valuation}
\label{subsec:kripke}

A Kripke model $\mathcal{K} = (W, R, V)$ has three parts.
\begin{itemize}[itemsep=1pt,topsep=2pt]
\item $W$ is a finite set of worlds. A world is a context in which every proposition has a truth value. Two worlds may disagree.
\item $R \subseteq W \times W$ is the accessibility relation. $R(w, w')$ means that $w'$ is one of the worlds consulted when a modal formula is evaluated at $w$ (Sect.~\ref{subsec:operators}).
\item $V$ is the valuation. $V(w, p)$ is the truth value of proposition $p$ in world $w$.
\end{itemize}
In this tutorial a world is always something concrete: a state of the whole system, such as which shelves of a warehouse are occupied (Sect.~\ref{sec:knowledge}); a sentence of a story (Sect.~\ref{sec:temporal}); the situation of one agent, such as a car at a crossing (Sect.~\ref{sec:deontic}); the events of a story that have happened (Sect.~\ref{sec:tom}); a hypothesis about a pull request, such as ``it has an operator-misuse bug'' (Sect.~\ref{sec:orchestration}); or a position or map cell that a drone may see (Sect.~\ref{sec:communication}).

\paragraph{Several agents.} With several agents there is one relation per agent, $\mathcal{K} = (W, \{R_i\}, V)$, and one pair of operators per agent, $\Box_i$ and $\Diamond_i$, which range over $R_i$. In epistemic logic $\Box_i\phi$ is written $K_i\phi$, ``agent $i$ knows $\phi$'', so $K_AK_Bp$ means ``$A$ knows that $B$ knows $p$''. Sects.~\ref{sec:knowledge}, \ref{sec:tom}, \ref{sec:orchestration} and~\ref{sec:communication} use this form; the running example and Sects.~\ref{sec:temporal}--\ref{sec:deontic} have a single relation.

\paragraph{Running example.} A car waits at a red light. The worlds are three situations it can be in at the next step: $a$, still waiting at the red light; $b$, driving off after the light has turned green; and $c$, driving through on red. Two propositions describe each situation, $\mathit{red}$ = ``the light is red'' and $\mathit{move}$ = ``the car moves'': both hold in $c$, only $\mathit{red}$ in $a$, only $\mathit{move}$ in $b$. The relation is deontic: $R(w,w')$ says that $w'$ is a permitted next situation after $w$. From $a$, waiting longer is permitted, driving off on green mostly, and driving through on red, which the rules forbid, still partly: $R(a,a) = 1$, $R(a,b) = 0.9$, $R(a,c) = 0.6$ (Fig.~\ref{fig:worked}). For now, take these numbers to mean ``yes'', ``mostly'' and ``partly''; Sect.~\ref{sec:method} makes them precise, and Sect.~\ref{sec:deontic} learns such a relation from the behaviour of compliant drivers. One global truth value for $\mathit{move}$ could not say that moving is wrong in $c$ but fine in $b$; with worlds, each situation has its own values, and $R$ says which situations count from where. The rule of the road is the formula $\Box(\mathit{red} \to \neg\mathit{move})$.

\subsection{Necessity and possibility}
\label{subsec:operators}

The two modal operators combine truth values across accessible worlds:
\begin{align}
\mathcal{K}, w \models \Box\phi \quad&\text{iff}\quad \phi \text{ holds in every } w' \text{ with } R(w,w'), \label{eq:box_classic}\\
\mathcal{K}, w \models \Diamond\phi \quad&\text{iff}\quad \phi \text{ holds in some } w' \text{ with } R(w,w'). \label{eq:dia_classic}
\end{align}
$\Box$ (``necessarily'') is a universal quantifier over the accessible worlds and $\Diamond$ (``possibly'') an existential one. The operators always mean this; what changes from one application to another is what the worlds and the relation stand for. Table~\ref{tab:modalities} lists the four readings used in this tutorial. In the running example, $\Box(\mathit{red}\to\neg\mathit{move})$ at $a$ states that every permitted next situation obeys the rule, that is, that stopping at red is obligatory; $\Diamond\mathit{move}$ at $a$ states that some permitted next situation has the car moving, that is, that moving on is permitted.

\begin{table}[t]
\centering
\caption{The four readings of the same two operators. ``Epistemic'' refers to knowledge, ``doxastic'' to belief, ``deontic'' to obligation and permission. Sects.~\ref{sec:knowledge}--\ref{sec:tom} each focus on one reading; the applications of Sects.~\ref{sec:orchestration}--\ref{sec:communication} combine them.}
\label{tab:modalities}
\small
\setlength{\tabcolsep}{4pt}
\begin{tabular}{@{}l>{\raggedright\arraybackslash}p{0.2\linewidth}>{\raggedright\arraybackslash}p{0.22\linewidth}>{\raggedright\arraybackslash}p{0.2\linewidth}>{\raggedright\arraybackslash}p{0.17\linewidth}@{}}
\toprule
Reading & $\Box\phi$ at $w$ means & $\Diamond\phi$ at $w$ means & $R(w,w')$ means & Used in \\
\midrule
Epistemic & $\phi$ is known at $w$ & $\phi$ is compatible with what is known at $w$ & $w'$ cannot be told apart from $w$ & Sects.~\ref{sec:knowledge}, \ref{sec:orchestration}, \ref{sec:communication}\\
Doxastic & $\phi$ is believed at $w$ & $\phi$ is compatible with what is believed at $w$ & $w'$ is consistent with the beliefs held at $w$ & Sects.~\ref{sec:knowledge}, \ref{sec:tom}, \ref{sec:communication}\\
Temporal & $\phi$ holds at every next step & $\phi$ holds at some next step & $w'$ can come after $w$ & Sect.~\ref{sec:temporal}\\
Deontic & $\phi$ is obligatory at $w$ & $\phi$ is permitted at $w$ & $w'$ is an acceptable successor of $w$ & Sect.~\ref{sec:deontic}; the running example\\
\bottomrule
\end{tabular}
\end{table}

\subsection{Frame axioms}
\label{subsec:frame_axioms}\label{sec:frame_axioms}

Different readings need relations with different shapes. These shapes are described by frame axioms, which are conditions on $R$ alone (Table~\ref{tab:frame_axioms}). Each condition on $R$ corresponds to a formula that becomes valid when the condition holds. For example, if every world is accessible from itself (reflexivity, axiom T), then $\Box\phi \to \phi$ is valid: what is known is true. Knowledge is usually modelled with T, 4 and 5 (the system S5); belief drops T, because a belief can be false, and adds D (the system KD45)~\citep{fagin2003reasoning}. K45 is KD45 without D. In this tutorial the frame axioms are not imposed during training, with the exceptions noted in each case study. Instead we measure after training how well the learned relation satisfies each of them (Sect.~\ref{subsec:audit}).

\begin{table}[t]
\centering
\caption{Frame axioms: the condition on $R$ and the formula it makes valid.}
\label{tab:frame_axioms}
\small
\setlength{\tabcolsep}{5pt}
\begin{tabular}{@{}lll>{\raggedright\arraybackslash}p{0.3\linewidth}@{}}
\toprule
Axiom & Condition on $R$ & Valid formula & Informal reading \\
\midrule
T (reflexive)  & $R(u,u)$ for all $u$                            & $\Box\phi \to \phi$                 & what is known is true \\
4 (transitive) & $R(u,v)\wedge R(v,w) \Rightarrow R(u,w)$       & $\Box\phi \to \Box\Box\phi$         & knowing implies knowing that one knows \\
B (symmetric)  & $R(u,v) \Rightarrow R(v,u)$                     & $\phi \to \Box\Diamond\phi$         & access goes both ways \\
5 (Euclidean)  & $R(u,v)\wedge R(u,w) \Rightarrow R(v,w)$       & $\Diamond\phi \to \Box\Diamond\phi$ & not knowing implies knowing that one does not know \\
D (serial)     & every $u$ has some $v$ with $R(u,v)$            & $\Box\phi \to \Diamond\phi$         & there is always a next world \\
\bottomrule
\end{tabular}
\end{table}

\section{Differentiable Modal Logic}
\label{sec:method}\label{sec:architecture}

This section describes the Modal Logic Neural Network (MLNN) of~\citet{sulc2026modal} in the form used in the rest of the tutorial, as implemented in the \tm{} library. There are four steps: replace true/false by numbers in $[0,1]$, make the relation a parameter, replace $\forall$ and $\exists$ by smooth minimum and maximum, and turn an axiom into a loss.

\subsection{Truth values in $[0,1]$}
\label{subsec:truth}

A truth value is a number in $[0,1]$: $1$ is true, $0$ is false, values in between are degrees of truth. The connectives are those of Łukasiewicz logic:
\begin{equation}
\neg a = 1-a,\quad\; a\wedge b = \max(0,\,a+b-1),\quad\; a\vee b = \min(1,\,a+b),\quad\; a\to b = \min(1,\,1-a+b).
\label{eq:luk}
\end{equation}
On the values $0$ and $1$ they agree with classical logic. The implication $a \to b$ equals $1$ whenever $b \ge a$ and drops linearly when $a$ exceeds $b$; this is what makes it useful as a loss (Sect.~\ref{subsec:loss}).

\paragraph{Lower and upper bounds.} Some evidence sources give two numbers about a proposition; a natural language inference model, for example, returns the probabilities that a sentence entails and that it contradicts a statement. A truth value may therefore be an interval $[L, U]$: $L$ is the degree to which $p$ is supported, $U$ is one minus the degree to which it is ruled out, and a single value is the case $L=U$. A source that both supports and rules out $p$ strongly gives $L > U$, an empty interval, which Sect.~\ref{subsec:loss} also counts as a contradiction (Eq.~\ref{eq:contra_bound}). \tm{} stores every truth value as such a pair.

\subsection{A learnable accessibility relation}
\label{subsec:accessibility}

The relation becomes a matrix $A_\theta \in [0,1]^{|W|\times|W|}$, where $A_\theta[w,w']$ is the degree to which $w'$ is accessible from $w$. We use two parameterizations.
\begin{itemize}[itemsep=1pt,topsep=2pt]
\item Dense: $A_\theta = \sigma(\widetilde{A})$ with a free logit matrix $\widetilde{A}\in\mathbb{R}^{|W|\times|W|}$ and the logistic function $\sigma$. This has $|W|^2$ parameters and only works for a fixed, small set of worlds. Each entry is learned separately.
\item Metric: each world $w$ has a feature vector $x_w$ (for example a sensor reading), an encoder $f_\theta$ maps it to an embedding, and $A_\theta[w,w'] = \sigma\bigl(s\, f_\theta(x_w)^\top g_\theta(x_{w'}) + b\bigr)$, with a second encoder $g_\theta$ for the target world and a learned scale $s$ and bias $b$. The number of parameters does not depend on $|W|$, and a new world gets an accessibility value as soon as its features are known. Using one encoder for both sides ($g_\theta = f_\theta$) makes the relation symmetric by construction; a directed relation needs two.
\end{itemize}
In both cases computing all entries costs $O(|W|^2)$ time and memory.

\subsection{Soft necessity and possibility}
\label{subsec:soft_ops}

Equations~\ref{eq:box_classic}--\ref{eq:dia_classic} use ``for every'' and ``for some''. With truth values in $[0,1]$ these become a minimum and a maximum over the accessible worlds, with the accessibility entering through the Łukasiewicz connectives: $\Box\phi(w) = \min_{w'} \bigl(A_\theta[w,w'] \to \phi(w')\bigr)$ and $\Diamond\phi(w) = \max_{w'} \bigl(A_\theta[w,w'] \wedge \phi(w')\bigr)$. A world with $A_\theta[w,w'] = 0$ contributes the value $1$ to the minimum and $0$ to the maximum, so it has no effect. A world with $A_\theta[w,w'] = 1$ contributes its own truth value. To obtain gradients that reach every term, the minimum and maximum are replaced by the log-sum-exp approximations
\begin{equation}
\smin(\mathbf{x}) = -\tau \log \sum_i e^{-x_i/\tau},\qquad
\smax(\mathbf{x}) = \tau \log \sum_i e^{x_i/\tau},
\label{eq:smin}
\end{equation}
with a temperature $\tau > 0$. For $n$ terms they satisfy $\min(\mathbf{x}) - \tau\log n \le \smin(\mathbf{x}) \le \min(\mathbf{x})$ and $\max(\mathbf{x}) \le \smax(\mathbf{x}) \le \max(\mathbf{x}) + \tau \log n$, so they approach the hard values as $\tau\to 0$, and the soft necessity never claims more than the hard one. The operators are
\begin{align}
\Box\phi(w) &= \mathrm{clip}_{[0,1]}\Bigl(\mathrm{smin}_{\tau,w'}\bigl[\,1 - A_\theta[w,w'] + \phi(w')\,\bigr]\Bigr), \label{eq:necessity}\\
\Diamond\phi(w) &= \mathrm{clip}_{[0,1]}\Bigl(\mathrm{smax}_{\tau,w'}\bigl[\,A_\theta[w,w'] + \phi(w') - 1\,\bigr]\Bigr). \label{eq:possibility}
\end{align}
The clip keeps the result in $[0,1]$. The two operators are dual, $\Diamond\phi = 1 - \Box(1-\phi)$, for every $\tau$: since $\smax(\mathbf{x}) = -\smin(-\mathbf{x})$, $1 - \smin\bigl[(1-A)+(1-\phi)\bigr] = \smax\bigl[A+\phi-1\bigr]$, and clipping to $[0,1]$ preserves the identity, so an implementation needs only one of them. We use $\tau = 0.1$ unless stated otherwise.

\paragraph{Interval bounds.} For interval truth values $[L, U]$ (Sect.~\ref{subsec:truth}) each operator returns an interval. For $\Box$, the lower bound is Eq.~\ref{eq:necessity} applied to $L_\phi$. The upper bound is a weighted average of the same terms with $U_\phi$ in their place,
\[
U_{\Box\phi}(w) = \textstyle\sum_{w'} \pi_{w'}\, x_{w'}, \qquad x_{w'} = 1 - A_\theta[w,w'] + U_\phi(w'), \qquad \pi = \mathrm{softmax}(-\mathbf{x}/\tau),
\]
which puts most weight on the smallest terms. An average is never below the minimum, so the interval contains the hard value~\citep{sulc2026modal}. For $\Diamond$ the roles swap: Eq.~\ref{eq:possibility} applied to $U_\phi$ gives the upper bound, and the average of $A_\theta + L_\phi - 1$ with weights $\mathrm{softmax}(\mathbf{x}/\tau)$ gives the lower bound. For a single truth value ($L_\phi = U_\phi$) the two ends differ by at most $\tau\log n$. The end used in a loss matters, as Sects.~\ref{sec:deontic}--\ref{sec:orchestration} show.

\begin{figure}[t]
\centering
\begin{tikzpicture}[font=\small, >=Stealth,
  world/.style={circle, draw=teal700, line width=0.8pt, fill=teal50, minimum size=9mm},
  edge/.style={->, line width=0.9pt, draw=slate700}]
\node[world] (a) at (0,0) {$a$};
\node[world] (b) at (2.7,0.9) {$b$};
\node[world] (c) at (2.7,-0.9) {$c$};
\node[anchor=north, text=slate700, align=center, font=\footnotesize] at (a.south) {waits at red\\$\mathit{red}{=}1,\ \mathit{move}{=}0$};
\node[anchor=west, text=slate700, align=left, font=\footnotesize] at (b.east) {drives off on green\\$\mathit{red}{=}0,\ \mathit{move}{=}1$};
\node[anchor=west, text=slate700, align=left, font=\footnotesize] at (c.east) {drives through on red\\$\mathit{red}{=}1,\ \mathit{move}{=}1$};
\draw[edge] (a) -- node[above, sloped] {$0.9$} (b);
\draw[edge] (a) -- node[below, sloped] {$0.6$} (c);
\draw[edge] (a) edge[loop above, looseness=6] node[above] {$1.0$} (a);
\node[anchor=west] at (6.1,0) {%
\begin{tabular}{@{}lccc@{}}
\toprule
from $a$ to & $a$ & $b$ & $c$\\
\midrule
$A[a,\cdot]$ & 1.0 & 0.9 & 0.6\\
$\phi = \mathit{red}\to\neg\mathit{move}$ & 1 & 1 & 0\\
$1-A+\phi$ (for $\Box\phi$) & 1.0 & 1.1 & \textbf{0.4}\\
$A+\mathit{move}-1$ (for $\Diamond\mathit{move}$) & 0.0 & \textbf{0.9} & 0.6\\
\midrule
\multicolumn{4}{@{}l}{$\Box\phi(a) = \smin = 0.400$ \quad(hard min $0.40$)}\\
\multicolumn{4}{@{}l}{$\Diamond\mathit{move}(a) = \smax = 0.905$ \quad(hard max $0.90$)}\\
\bottomrule
\end{tabular}};
\end{tikzpicture}
\caption{The running example. Left: the situation $a$ of a car waiting at a red light and its possible next situations, with the permission $A$ on the arrows and the two propositions of each situation. Right: the terms of Eqs.~\ref{eq:necessity}--\ref{eq:possibility} at $a$ ($\tau=0.1$). Driving through on red, still permitted at $0.6$, determines $\Box(\mathit{red}\to\neg\mathit{move})(a)$; driving off on green determines $\Diamond\mathit{move}(a)$. The numbers are computed with \tm{}.}
\label{fig:worked}
\end{figure}

\paragraph{Example.} Figure~\ref{fig:worked} evaluates both operators at $a$. The rule $\Box(\mathit{red}\to\neg\mathit{move})$ states that every permitted next situation obeys it. Waiting and driving off on green obey it, so they do not lower its value. Driving through on red breaks it and is still permitted at $0.6$, so the rule holds only to $1 - 0.6 = 0.40$. The permission $\Diamond\mathit{move}$ states that some permitted next situation has the car moving. Driving off on green does, and it is permitted at $0.9$, so moving is permitted to $0.90$. If driving through on red were forbidden ($A[a,c] = 0$), the rule would hold fully, and moving would still be permitted, on green. The numbers on the arrows decide which situations count.

\subsection{Axioms and the contradiction loss}
\label{subsec:loss}

\paragraph{Axioms.} Formulas are built from propositions with the connectives of Eq.~\ref{eq:luk} and the modal operators:
\begin{equation}
\phi \;::=\; p \;\mid\; \neg\phi \;\mid\; \phi\wedge\phi \;\mid\; \phi\vee\phi \;\mid\; \phi\to\phi \;\mid\; \Box\phi \;\mid\; \Diamond\phi \;\mid\; \phi\,\mathcal{U}\,\phi .
\label{eq:grammar}
\end{equation}
$\phi\,\mathcal{U}\,\psi$ (``$\phi$ until $\psi$'') is used only in Sect.~\ref{sec:temporal} and is defined there. An axiom is a formula that we assert to be true in every world of a chosen set. We use two kinds. Frame axioms (Table~\ref{tab:frame_axioms}) are conditions on the relation alone; we measure them after training. Domain axioms describe the application, for example the traffic rule $\Box(\mathit{red}\to\neg\mathit{move})$. Domain axioms are what we train on. Each case study states its domain axiom as a formula of Eq.~\ref{eq:grammar}.

\paragraph{Loss.} Asserting an axiom $\alpha$ at world $w$ costs $1 - \alpha(w)$, its distance from being true. For an axiom of the form $\phi \to \psi$ this cost is
\begin{equation}
1 - \min(1,\,1-\phi+\psi) \;=\; \relu(\phi - \psi),
\label{eq:contra}
\end{equation}
which is zero when $\psi \ge \phi$ and grows linearly with the gap. When $\alpha(w)$ is an interval $[L, U]$, asserting it true costs $1 - U$ and asserting it false costs $L$, the least violation the interval allows: this is the bound inversion of Eq.~\ref{eq:contra_bound} below, with the assertion pinned to $[1,1]$ or $[0,0]$. Sects.~\ref{sec:deontic} and~\ref{sec:orchestration} use this cost; Sect.~\ref{sec:tom} explains why it uses the midpoint of the bounds instead. For an interval there is also a second cost,
\begin{equation}
\relu(L - U),
\label{eq:contra_bound}
\end{equation}
which is positive exactly when the interval is empty. We call both the contradiction loss $\Lcontra$; \tm{} computes either with \textsf{F.contradiction}. The total loss sums (or averages) these costs over the worlds where the axiom is asserted.

\paragraph{Example (continued).} Now we train on the rule: we assert $\Box(\mathit{red}\to\neg\mathit{move})$ at $a$. It holds only to $0.40$, so asserting it costs $0.60$. The cost comes from driving through on red, so gradient descent lowers that permission, $A[a,c]$. Training stops when the loss reaches $0$, here at $A[a,c] \approx 0.03$ (the loss uses the upper end of the rule's value, which reaches $1$ first). The rule then holds to $0.97$, and moving on green is still permitted ($0.90$); Listing~\ref{lst:running} does all of this in \tm{}. Two points recur in the case studies. First, forbidding everything would also satisfy the rule; in Sect.~\ref{sec:deontic}, examples of compliant drivers and the axiom D keep the permitted moves up. Second, an entry that no axiom ever tests keeps its initial value, so a high entry can mean ``tested and never contradicted'' or ``never tested''.

\begin{figure}[t]
\centering
\resizebox{\textwidth}{!}{%
\begin{tikzpicture}[
    font=\sffamily, >=Stealth,
    world/.style={circle, draw=teal700, line width=0.9pt, fill=teal50,
        minimum size=1.25cm, text=teal700, font=\LARGE},
    val_box/.style={rectangle, draw=teal700, line width=0.6pt, fill=teal50!50,
        rounded corners=1.0mm, minimum height=0.7cm, font=\large, align=center},
    learn_strong/.style={->, line width=1.7pt, draw=rose700, shorten >=2pt, shorten <=2pt},
    learn_mid/.style={->, line width=1.05pt, draw=rose700!75, shorten >=2pt, shorten <=2pt},
    op_box/.style={rectangle, draw=slate700, fill=slate50, line width=0.6pt,
        rounded corners=1.5mm, align=center, inner sep=6pt, font=\large},
    ax_box/.style={op_box, draw=amber600, fill=amber50},
    loss_box/.style={op_box, draw=rose700, fill=rose700!6},
    flow/.style={->, line width=1pt, draw=slate500},
    grad/.style={->, line width=1.8pt, dashed, draw=amber600},
    panel_label/.style={font=\Large\bfseries, text=slate700},
    note/.style={font=\normalsize\itshape, text=slate700, align=center},
    cell/.style={rectangle, draw=white, line width=1pt, minimum width=1.25cm,
        minimum height=0.9cm, font=\large},
]
\node[panel_label] at (2.8, 4.75) {1\; Worlds and relation};
\node[world] (a) at (0.9, 2.0) {$a$};
\node[world] (b) at (4.6, 3.2) {$b$};
\node[world] (c) at (4.6, 0.6) {$c$};
\node[val_box, anchor=north, yshift=-2mm] at (a.south) {waits at red};
\node[val_box, anchor=west, xshift=2mm, font=\normalsize] at (b.east) {green,\\drives};
\node[val_box, anchor=west, xshift=2mm, font=\normalsize] at (c.east) {red,\\drives};
\draw[learn_strong] (a) -- node[above=1.5mm, sloped, font=\large, text=rose700] (ab) {$A_\theta{=}0.9$} (b);
\draw[learn_mid] (a) -- node[below=0.5mm, sloped, pos=0.72, font=\large, text=rose700!85] {$0.6$} (c);
\draw[learn_strong] (a) edge[in=110, out=70, looseness=7] node[above, font=\large, text=rose700] {$1.0$} (a);

\node[panel_label] at (9.9, 4.75) {2\; Formulas};
\node[op_box, minimum width=4.6cm] (ops) at (9.9, 3.0)
    {$\Diamond\mathit{move}(a) = 0.90$};
\node[ax_box, minimum width=4.6cm] (ax) at (9.9, 0.75)
    {axiom $\Box(\mathit{red}\to\neg\mathit{move})$ at $a$\\[3pt] $\alpha(a) = 0.40$};

\node[panel_label] at (16.4, 4.75) {3\; Loss};
\node[loss_box, minimum width=3.6cm] (loss) at (16.4, 1.9)
    {$\Lcontra = 1 - \alpha(a)$\\[3pt] $= 0.60$};

\node[panel_label] at (22.6, 4.75) {4\; Result};
\node[cell, fill=rose700!85, text=white] at (21.35, 2.6) {$1.00$};
\node[cell, fill=rose700!75, text=white] at (22.6, 2.6) {$0.90$};
\node[cell, fill=rose700!8, text=slate900] at (23.85, 2.6) {$0.03$};
\node[font=\normalsize, text=slate500] at (21.35, 1.9) {$a$};
\node[font=\normalsize, text=slate500] at (22.6, 1.9) {$b$};
\node[font=\normalsize, text=slate500] at (23.85, 1.9) {$c$};
\node[font=\large, text=slate700, align=center] at (22.6, 0.9) {$\Box(\mathit{red}\to\neg\mathit{move})(a) = 0.97$\\[2pt] $\Diamond\mathit{move}(a) = 0.90$};

\draw[flow] (7.05, 3.2) -- (ops.west);
\draw[flow] (7.05, 0.6) -- (ax.west);
\draw[flow] (ops.east) -- (loss.north west);
\draw[flow] (ax.east) -- (loss.south west);
\draw[flow] (loss.east) -- node[above, align=center, font=\normalsize\itshape, text=slate700] {after\\training} (20.55, 1.9);
\draw[grad] (loss.south) -- ++(0, -1.75) -| node[pos=0.27, below=2pt, font=\large, text=amber600]
    {gradient step:\; $A_\theta[a,c]\colon 0.6 \to 0.03$} (2.75, 1.25);
\end{tikzpicture}}
\caption{The training loop on the running example (numbers from \tm{}). (1) The worlds and the valuation $V$ are fixed before training; the relation $A_\theta$ is the only parameter. (2) The operators and the domain axiom are evaluated at $a$ with Eqs.~\ref{eq:necessity}--\ref{eq:possibility}; the axiom is the rule of the road, and driving through on red ($c$) breaks it. (3) The contradiction loss measures how far the axiom is from true, and its gradient lowers the permission to drive through on red until the loss is $0$. (4) After training, the relation and the operators are inspected: the rule holds at $a$ (hard value), and moving on green is still permitted.}
\label{fig:loop}
\end{figure}

\paragraph{Role of the modal operator.} If the axiom is $\Box(\phi\to\psi)$ asserted at every world, the loss is zero exactly when every edge satisfies its own implication, since in the hard limit $\Box$ is a minimum. In that limit the zero-loss set is therefore the same for $\Box$, a sum, or a mean over the edges (\citealp[Prop.~2]{sulc2026modal}, for one doxastic axiom); only the distribution of the gradient differs ($\smin$ concentrates it on the worst violation). The operators matter more when their result is used further: in Sects.~\ref{sec:deontic} and~\ref{sec:orchestration} $\Box$ and $\Diamond$ are the outputs, and in Sect.~\ref{sec:tom} $\Box$ is nested up to four times.

\subsection{The training loop}
\label{subsec:workflow}

Every case study is an instance of the same loop (Fig.~\ref{fig:loop}). Choose the worlds $W$ and the propositions, and compute the valuation $V$ from data (a simulator, the answers of other agents, a sentence encoder); $V$ is fixed during training unless stated. Choose a parameterization of $A_\theta$ (Sect.~\ref{subsec:accessibility}) and an initial value, which untested entries will keep. State the domain axiom and evaluate it at each world with Eqs.~\ref{eq:luk}--\ref{eq:possibility}. Minimize $\Lcontra$ (Eqs.~\ref{eq:contra}--\ref{eq:contra_bound}) over $\theta$ by gradient descent. Finally, inspect the entries of $A_\theta$, the values of $\Box$ and $\Diamond$ at each world, and check the frame axioms (Sect.~\ref{subsec:audit}).

\begin{table}[t]
\centering
\caption{The six case studies as instances of the training loop: the worlds, the learned relation, the axioms trained on, the operators used, and the simpler method that does as well, or nearly as well, on the task itself.}
\label{tab:studies}
\footnotesize
\setlength{\tabcolsep}{3pt}
\renewcommand{\arraystretch}{1.15}
\begin{tabular}{@{}>{\raggedright\arraybackslash}p{0.09\linewidth}>{\raggedright\arraybackslash}p{0.14\linewidth}>{\raggedright\arraybackslash}p{0.17\linewidth}>{\raggedright\arraybackslash}p{0.19\linewidth}>{\raggedright\arraybackslash}p{0.17\linewidth}>{\raggedright\arraybackslash}p{0.17\linewidth}@{}}
\toprule
Sect. & Worlds & $A_\theta[w,w']$ & Axioms & Operators & Does as well \\
\midrule
\ref{sec:knowledge} Epistemic & placements of boxes on shelves & robot $i$ cannot tell $w'$ from $w$ & each report: $K_i\ell$, or $\Diamond_i x_k \wedge \Diamond_i \neg x_k$ & $\Box$ and $\Diamond$ together; nested $K_iK_j$ & a classifier, on single-shelf reports \\
\ref{sec:temporal} Temporal & sentences of a story & $w'$ can follow $w$ & the true order scores highest & Until over the four steps & a position classifier, on pairs \\
\ref{sec:deontic} Deontic & situations of a car & $w'$ is a permitted next situation & demonstrations; $\Box\neg\mathit{conflict}$; $\Diamond\top$ (D) & $\Box$ for the rule, $\Diamond$ for seriality & a lookup, near the data \\
\ref{sec:tom} Doxastic & sets of story events & person $p$ considers $w'$ possible & first-order answers, through $\Box_p$ & $\Box$ nested up to four times & the exact rule, given who observed what \\
\ref{sec:orchestration} Orchestration & hypotheses about a PR; reports & reviewer $i$ may report $w'$ under $w$ & $\Diamond_i$(observed report) at the true hypothesis & $\Box$, $\Diamond$ over the hypotheses left & Dawid--Skene, logistic regression (more accurate) \\
\ref{sec:communication} Coordination & positions and map cells & drone $i$ would see $w'$ & ``I see $j$''; a belief $B_aK_b\phi$ does not contradict $K_a$ & $\Box$ over a drone's view; $K_aK_b$; $\Box$ / $\Diamond$ map & ranking by failed checks, nearly ($0.86$ vs $0.99$) \\
\bottomrule
\end{tabular}
\end{table}

\subsection{Checking the frame axioms}
\label{subsec:audit}

After training we check which frame axioms (Table~\ref{tab:frame_axioms}) the learned matrix $A$ satisfies. Following~\citet{sulc2026modal}, each axiom gets a score between $0$ and $1$, the average truth value of its condition:
\begin{align*}
S_T &= \operatorname*{avg}_u A_{uu}, &
S_4 &= \operatorname*{avg}_{u,v,w}\bigl((A_{uv}\wedge A_{vw})\to A_{uw}\bigr),\\
S_B &= \operatorname*{avg}_{u,v}\bigl(1 - |A_{uv}-A_{vu}|\bigr), &
S_5 &= \operatorname*{avg}_{u,v,w}\bigl((A_{uv}\wedge A_{uw})\to A_{vw}\bigr).
\end{align*}
Each satisfied axiom says something about the agent whose relation it is. In the epistemic reading, T means that every world is accessible from itself, so what the agent knows is true (a relation that fails T describes belief, which can be false); B that access goes both ways; 4 that a world reachable in two steps is reachable in one, so an agent that knows $\phi$ knows that it knows; and 5 that two worlds accessible from the same world are accessible from each other, so an agent that does not know $\phi$ knows that it does not know. The conditions of 4 and 5 hold trivially when the edges are weak, so Sects.~\ref{sec:knowledge} and~\ref{sec:deontic} score them only on the triples that test them (a fraction $\kappa$ of all triples) and compare with the shuffled null, the same score on a copy of $A$ whose off-diagonal entries are permuted within each row.\footnote{To build the shuffled copy, the off-diagonal entries of each row are put in random order while the diagonal stays fixed. Sect.~\ref{sec:knowledge} also keeps the largest off-diagonal entry of each row fixed, as in~\citet{sulc2026modal}, so the copy is closer to the learned matrix and harder to beat; Sect.~\ref{sec:deontic} uses the version in \tm{}, which fixes only the diagonal.}

\vspace{-1em}\section{Epistemic and Doxastic: What Each Robot Sees}\vspace{-0.25em}
\label{sec:knowledge}\label{sec:epistemic}

The first case study learns who knows what. Robots watch the shelves of a warehouse, each through its own camera, and we are told only what each robot says it knows, never which placements it can tell apart. Since ``I know'' is a $\Box$ formula and ``I do not know'' a $\Diamond$ formula, the relation can be learned only through the operators. Once learned, it answers questions that never appear in training, and checking the frame axioms (Sect.~\ref{subsec:audit}) separates knowledge (S5) from belief (KD45).

\paragraph{Setting.} A warehouse has eight shelves and four boxes, and every robot knows there are four. A world is one placement of the boxes, so there are $\binom{8}{4} = 70$ worlds, and $x_k$ means ``shelf $k$ is occupied''. The camera of robot $i$ sees the shelves in $O_i$: $O_1=\{1,2,3,4\}$, $O_2=\{3,4,5\}$, $O_3=\{5,6,7,8\}$, $O_4=\{2,7\}$. The model is not told these sets. Robot $i$ cannot tell apart two placements that look the same on its shelves, so its true relation $R_i(s,s')$ holds exactly when $s$ and $s'$ agree on $O_i$. This is an equivalence relation, the frame of S5 (Table~\ref{tab:frame_axioms}). We write $K_i\phi$, ``robot $i$ knows $\phi$'', for $\Box_i\phi$: $K_i\phi$ holds at $s$ when $\phi$ holds in every placement robot $i$ cannot tell apart from $s$. Whichever of them is the true one, $\phi$ is true, so the robot can be sure of it. Because the total is known, a robot can know more than it sees. If robot~2 sees no box on shelves 3--5, the four boxes sit on four of the other five shelves, so of any two of these at least one is occupied: robot 2 knows $x_a \vee x_b$ for every such pair, but not $x_a$ alone (Fig.~\ref{fig:knowledge_scenario}).

\begin{figure}[!b]
\centering
\begin{tikzpicture}[font=\small, >=Stealth, x=1cm, y=1cm]
\foreach \k in {1,...,8} {
  \draw[slate700, line width=0.6pt, fill=white] ({0.85*\k},0) rectangle ++(0.7,0.55);
  \node[font=\scriptsize, text=slate500] at ({0.85*\k+0.35},-0.22) {$x_{\k}$};
}
\foreach \k in {1,3,4,7} \fill[amber600] ({0.85*\k+0.15},0.08) rectangle ++(0.4,0.39);
\node[font=\scriptsize, text=slate500, anchor=west] at (7.7,0.28) {world $s$: $x(s) = (1,0,1,1,0,0,1,0)$};
\foreach \i/\a/\b/\y/\c in {1/1/4/0.95/teal700, 3/5/8/0.95/teal700, 2/3/5/1.4/slate700} {
  \draw[\c, line width=1pt] ({0.85*\a},\y) -- ({0.85*\b+0.7},\y);
  \draw[\c, line width=1pt] ({0.85*\a},\y) -- ++(0,-0.13);
  \draw[\c, line width=1pt] ({0.85*\b+0.7},\y) -- ++(0,-0.13);
  \node[font=\scriptsize, text=\c, above] at ({0.85*(\a+\b)/2+0.35},\y) {robot \i{} sees $O_{\i}$};
}
\foreach \a in {2,7} {
  \draw[amber600, line width=1pt] ({0.85*\a},1.85) -- ++(0.7,0);
  \draw[amber600, line width=1pt] ({0.85*\a},1.85) -- ++(0,-0.13);
  \draw[amber600, line width=1pt] ({0.85*\a+0.7},1.85) -- ++(0,-0.13);
}
\node[font=\scriptsize, text=amber600, above] at (4.5,1.85) {robot 4 sees $O_4 = \{2, 7\}$};
\begin{scope}[yshift=-1.0cm]
\foreach \lab/\row/\yy in {s/{1,0,1,1,0,0,1,0}/0, s'/{0,1,1,1,0,1,0,0}/-0.45} {
  \node[font=\scriptsize, anchor=east] at (0.75,\yy+0.17) {$\lab$};
  \foreach \v [count=\k] in \row {
    \ifnum\v=1 \fill[amber600] ({0.85+0.42*(\k-1)},\yy) rectangle ++(0.34,0.34);
    \else \draw[slate500, line width=0.4pt] ({0.85+0.42*(\k-1)},\yy) rectangle ++(0.34,0.34); \fi
  }
}
\draw[slate700, line width=0.9pt] ({0.85+0.42*2-0.05},-0.5) rectangle ({0.85+0.42*4+0.39},0.39);
\node[font=\scriptsize, text=slate700, anchor=north] at ({0.85+0.42*3.5},-0.52) {$O_2$};
\node[font=\scriptsize, anchor=west, align=left, text width=4.6cm] at (4.5,-0.1) {$s$ and $s'$ agree on $O_2$, so robot~2 cannot tell them apart: $R_2(s,s')$ holds.\\ They differ on shelf 1, so $R_1(s,s')$ does not.};
\end{scope}
\node[draw=slate700, fill=slate50, rounded corners=3pt, inner sep=4pt, align=left, font=\scriptsize, anchor=north west, text width=5.2cm] at (9.8,-0.75) {\textbf{robot 2 in world $s$ says}\\ ``I know shelf 3 is occupied'' \hfill $K_2 x_3$\\ ``I know shelf 5 is empty'' \hfill $K_2 \neg x_5$\\ ``I do not know shelf 1'' \hfill $\Diamond_2 x_1 \wedge \Diamond_2 \neg x_1$};
\end{tikzpicture}
\caption{The knowledge scenario. Top: eight shelves, four boxes, and the shelves each robot's camera covers; the camera sets $O_i$ are hidden from the model. Bottom left: two placements that robot~2 cannot tell apart. Bottom right: the only data the model sees are such statements, one per robot and shelf in the reported worlds; no pair $(s, s')$ is ever labelled.}
\label{fig:knowledge_scenario}
\end{figure}
\paragraph{Data.} In a world $s$, each robot states for each shelf $k$ ``I know shelf $k$ is occupied'' ($K_i x_k$), ``I know it is empty'' ($K_i\neg x_k$), or ``I do not know'' ($\neg K_i x_k \wedge \neg K_i\neg x_k$, which equals $\Diamond_i x_k \wedge \Diamond_i\neg x_k$). We give the reports of $28$ of the $70$ worlds ($40\%$) and hold out the rest.

\begin{glance}{Setup}
Worlds & The $70$ placements of $4$ boxes on $8$ shelves. \\
Propositions & $x_k$ = ``shelf $k$ is occupied'' ($0$ or $1$). \\
Relation & $A_i[s,s']$ = robot $i$ cannot tell $s'$ from $s$; one matrix per robot. \\
 & Metric form ($36$ parameters): $\sigma\bigl(\beta_i - \sum_k \mathrm{softplus}(w_{ik})\,[x_k(s)\neq x_k(s')]\bigr)$. \\
 & Dense form: $\sigma(\widetilde{A}_i)$, initial value $0.02$. \\
Axioms & Each report, as a formula at its world; no pair of worlds is labelled. \\
Loss & ``Know'' report of a literal $\ell$ ($x_k$ or $\neg x_k$): $1-\Box_i\ell(s)$. \\
 & ``Do not know'' report: $\bigl(1-\Diamond_i x_k(s)\bigr) + \bigl(1-\Diamond_i\neg x_k(s)\bigr)$; all terms averaged. \\
Output & Shelf weights $w_{ik}$; new formulas on $A_i$ thresholded at $0.5$; frame axioms. \\
\end{glance}

\paragraph{Both operators are needed.} A ``know'' report is a $\Box$ formula, and the empty relation satisfies every such formula: a robot that considers no world possible knows everything. A ``do not know'' report is a $\Diamond$ formula; it needs a witness, a world the robot cannot rule out in which the shelf has the other value, and the full relation satisfies every such formula. Only the two together determine the relation: trained on either kind alone, the relation becomes empty or full.

\paragraph{Two parameterizations.} The dense form learns each entry on its own. The metric form gives robot $i$ one weight $w_{ik}$ per shelf: a large weight means that a difference on shelf $k$ makes a world inaccessible, i.e.\ the robot sees shelf $k$; we count it as seen when $\mathrm{softplus}(w_{ik}) > \beta_i$. The weights start at $\mathrm{softplus}(w_{ik}) = 2$. Started instead between $0.5$ and $1.0$, $20$--$30\%$ of the runs stop at a wrong camera set, but with a training loss two to five times higher, so the start can be chosen without knowing the answer. We lower $\tau$ from $0.1$ to $0.02$ during training: over $70$ worlds the soft minimum at $\tau = 0.1$ lies up to $\tau\log 70 = 0.42$ below the hard one, so a true $\Box$ formula could not reach $1$.

\begin{figure}[t]
\centering
\includegraphics[width=\linewidth]{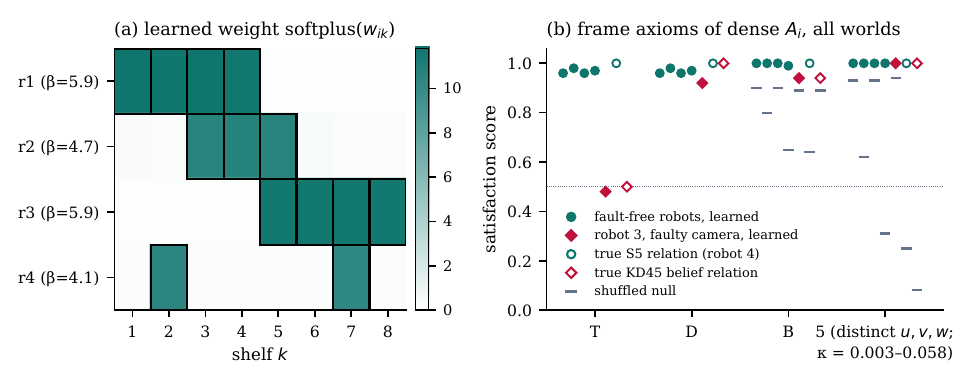}
\caption{Learning what each robot sees. (a) Metric relation, $40\%$ of worlds reported: learned shelf weights $\mathrm{softplus}(w_{ik})$ (mean over five seeds); black boxes mark the true camera sets $O_i$. (b) Frame-axiom scores of the dense relations trained on all 70 worlds, diagonal free, no prior. The four fault-free robots (teal) satisfy T, D, B and 5; robot 3, its camera stuck on shelf 7 (crimson), fails T and keeps D, B and 5. Hollow markers: the true relations (S5 for robot 4, KD45 for robot 3). Grey ticks: shuffled null. For axiom 5 the score is taken on the triples with distinct $u,v,w$ that test it; their fraction $\kappa$ is under the axis. Axiom 4 is not plotted: on the triples that test it every relation shown scores $1.00$, above its null.}
\label{fig:knowledge_relation}
\end{figure}

\paragraph{Results.} The metric form recovers every camera set $O_i$ exactly in all five seeds (Fig.~\ref{fig:knowledge_relation}a) and reproduces every report in the held-out worlds. The dense form fails there: no report constrains the row of a held-out world, so the row keeps its initial value $0.02$ and the robot appears to know everything, the untested entries of Sect.~\ref{subsec:loss}. For robot~2 (seed $42$), the dense relation recovers the true row in $25$ of the $28$ reported worlds and in none of the other $42$; the metric relation recovers all $70$. On the reports themselves two non-modal baselines do as well: a logistic regression per robot and shelf on the eight shelf bits reproduces $99\%$ of the held-out reports, and a hand-built Kripke model that takes as $O_i$ the shelves reported as known in every reported world is exact. The metric form works because its shelf-by-shelf structure matches how the data were made.

\paragraph{New questions.} Once $A_i$ is learned, any formula can be evaluated on it. The metric model answers four kinds of questions that never appear in the data, all exactly: $K_i(x_a\vee x_b)$, $K_i(x_a\wedge x_b)$, $K_iK_jx_k$ (``robot $i$ knows that robot $j$ knows shelf $k$'') and $K_ix_k\wedge K_jx_k$. A classifier of single-shelf reports has to combine its answers by rules such as $K(a\vee b) := Ka\vee Kb$, and scores $0.03$--$0.05$ lower. That rule is wrong even with perfect single-shelf answers, because knowledge does not distribute over disjunction: robot~2 above knows $x_a\vee x_b$ without knowing either shelf.

\paragraph{Frame axioms as a prior.} The frame axioms of S5 can be added to the dense form as soft penalties whose zero set is the relations that satisfy them~\citep[Thm.~3]{sulc2026modal}; for symmetry we use $R_B=\mathrm{avg}_{u,v}(A_{uv}-A_{vu})^2$, the paper's $\|A-A^\top\|_F^2$ averaged instead of summed, with weight $1$; they copy what a reported world has learned to the unreported worlds of its class. With $10\%$ of the worlds reported, held-out accuracy of the dense form rises from $0.21$ to $0.91$, and to $0.97$ at $40\%$.

\paragraph{A faulty camera.} Now robot~3's camera is stuck at ``occupied'' on shelf~7. In the $35$ worlds where shelf~7 is empty, robot~3 still says ``I know shelf 7 is occupied'', and in four of them it also wrongly says it knows that a shelf among 1--4 is empty. Statements that can be false are beliefs, not knowledge, so robot~3's true relation fails T and is KD45. We train the dense form on all $70$ worlds, with no axiom imposed, and score the frame axioms (Fig.~\ref{fig:knowledge_relation}b). The fault-free robots satisfy T ($S_T = 0.96$--$0.98$), B and 5. Robot~3 fails only T ($S_T = 0.48$, against $0.50$ for its true relation): to say ``I know shelf 7 is occupied'' in a world where it is empty, the robot must not consider that world possible.

\begin{table}[t]
\centering
\caption{Reports of the faulty camera, per robot and shelf: reports that no relation can satisfy when T is imposed / ``I know'' reports that are false in the actual world. The two counts coincide.}
\label{tab:faulty_reports}
\small
\setlength{\tabcolsep}{5pt}
\begin{tabular}{@{}lcccccccc@{}}
\toprule
Shelf & 1 & 2 & 3 & 4 & 5 & 6 & 7 & 8 \\
\midrule
Robot 1 & 0 / 0 & 0 / 0 & 0 / 0 & 0 / 0 & 0 / 0 & 0 / 0 & 0 / 0 & 0 / 0 \\
Robot 2 & 0 / 0 & 0 / 0 & 0 / 0 & 0 / 0 & 0 / 0 & 0 / 0 & 0 / 0 & 0 / 0 \\
Robot 3 & 1 / 1 & 1 / 1 & 1 / 1 & 1 / 1 & 0 / 0 & 0 / 0 & 35 / 35 & 0 / 0 \\
Robot 4 & 0 / 0 & 0 / 0 & 0 / 0 & 0 / 0 & 0 / 0 & 0 / 0 & 0 / 0 & 0 / 0 \\
\bottomrule
\end{tabular}
\end{table}

\paragraph{Imposing axiom T.} Axiom T can also be imposed during training. Fixed exactly, with the diagonal set to $1$, it leaves robot~3's $39$ false ``I know'' reports violated (Table~\ref{tab:faulty_reports}), so the violations point at the fault. Imposed softly, as S5 penalties, it hides the fault: robot~3's $S_T$ rises to $0.98$, as if its camera worked, and only a slightly worse fit to its own reports remains ($0.91$ instead of $0.96$). Penalties for 4 and 5 alone (K45) fit robot~3 with no violation and leave $S_T$ at $0.49$. An imposed axiom can no longer be checked, so impose only what is known to hold; where belief is possible, use K45.

\paragraph{Summary.} The relation was learned from statements alone, without any labelled pair of worlds, and both $\Box$ and $\Diamond$ were needed. Once learned, it answers new and nested questions. The axioms of S5 hold for the fault-free robots; for the faulty one only T fails. On single-shelf reports a classifier does as well, and a hand-built Kripke model with the right assumption is equally exact.

\vspace{-1em}\section{Temporal: Ordering the Sentences of a Story}\vspace{-0.25em}
\label{sec:temporal}

In this case study the worlds are the five sentences of a story, and the relation $A_\theta[i,j]$ says that sentence $s_j$ can follow sentence $s_i$. Model checking and runtime verification are given the transitions, or the observed runs, and check a formula~(\citealp{baier2008principles}; \citealp{bauer2011rv}); here the formula is fixed and the relation is learned from stories, so that one Until formula ranks the true order of a scrambled story above every other order. This is the sentence-ordering experiment of \citet{sulc2026modal}, with a small fix to the score that changes its numbers slightly.

\begin{glance}{Setup}
Worlds & The five sentences of a story. \\
 & Each is embedded as $h_t\in\mathbb{R}^{384}$ by a frozen sentence encoder, \\
 & \textsf{all-MiniLM-L6-v2}~\citep{reimers2019sentence}. \\
Relation & $A_\theta[i,j]$ = sentence $s_j$ can follow sentence $s_i$. \\
 & Metric form with two linear encoders (Eq.~\ref{eq:order_C}), $24{,}578$ parameters. \\
Formula & ``Every step of $\sigma$ is coherent until the last sentence'' (Until, Eq.~\ref{eq:order_score}). \\
Loss & The true order must score at least $m = 0.1$ above each of the other $119$ orders: \\
 & the contradiction loss of that implication (Eq.~\ref{eq:order_loss}). \\
Output & The best of the $120$ orders, and the $5\times5$ relation behind it. \\
Baselines & The mean cosine of adjacent sentences. \\
 & A classifier of each sentence's position, with no relation between sentences. \\
\end{glance}

\paragraph{Data.} $98\,155$ deduplicated stories from ROCStories Spring 2016 and Winter 2017~\citep{mostafazadeh2016roc}, of which $2\,000$ are held out for in-domain validation; the $1\,571$ Story Cloze 2018 validation stories~\citep{sharma2018tackling} (with their correct ending) are an out-of-distribution test, disjoint from the training pool. Each story is shown in a random scramble, and all $120$ orders are scored. Similarity alone cannot order the sentences, which share characters, vocabulary and tone: ordering by the mean cosine of adjacent sentence embeddings recovers $0.008$ of the stories exactly, chance level ($1/120$). The signal is directional (``then'', tense, cause before effect), which is what a temporal relation expresses.

\paragraph{Relation.} The relation is the metric form of Sect.~\ref{subsec:accessibility} with two encoders, here two linear maps:
\begin{equation}
A_\theta[i, j] \;=\; \sigma\!\bigl(\alpha\,\langle P_{\!\text{src}}\, h_i,\; P_{\!\text{tgt}}\, h_j\rangle + \beta\bigr),
\qquad P_{\!\text{src}}, P_{\!\text{tgt}} \in \mathbb{R}^{32 \times 384},
\label{eq:order_C}
\end{equation}
with learnable scalars $\alpha, \beta$. Because $P_{\!\text{src}} \neq P_{\!\text{tgt}}$, the relation can tell ``$i$ then $j$'' from ``$j$ then $i$''.

\begin{figure}[t]
\centering
\input{i/agent_debug_temporal_ordering/ordering_interpretability_tikz}
\caption{\textbf{The minimal model on four ROCStories examples} (seed $42$). Top: each ``$w_t\,\#k$'' token gives the true position $k$ of the sentence at scrambled position $w_t$, so a recovered chain is correct if it lists $\#1,\dots,\#5$ in order. The left cell gives the verdict, the recovered chain $\sigma^\star$ listed by true position, and the score margin $\Delta = \text{Score}(\sigma^\star) - \text{Score}(\sigma_{\text{true}})$. Bottom: the learned $A_\theta$ of each example (rose low, teal high); thick black border: the chain $\sigma^\star$; dashed amber: the true chain where it differs.}
\label{fig:order_interpretability}
\end{figure}

\paragraph{Score.} A candidate order $\sigma$ visits the five sentences one after another, in four steps from $\sigma(t)$ to $\sigma(t+1)$, $t = 0,\dots,3$. The truth of ``step $t$ is coherent'' is the relation's entry for that step, $\phi_t = A_\theta[\sigma(t), \sigma(t+1)]$. A good order is coherent at every step until it reaches the last sentence: with $\psi$ = ``this is the last sentence'', this is the Until formula $\phi\,\mathcal{U}\,\psi$, ``$\phi$ holds until $\psi$ does''. Evaluated at the first sentence of a five-sentence chain, it holds when all four steps are coherent, so in Łukasiewicz logic it is the conjunction of the four step values,
\begin{equation}
\text{Score}(\sigma) \;=\; (\phi\;\mathcal{U}\;\psi)\,(w_0) \;=\; \max\Bigl(0,\; \textstyle\sum_{t=0}^{3} A_\theta[\sigma(t), \sigma(t+1)] - 3\Bigr),
\label{eq:order_score}
\end{equation}
computed with \textsf{torchmodal.functional.until}. The score is $1$ when every step has $A_\theta = 1$, drops by the total shortfall $\sum_t (1 - \phi_t)$, and is $0$ once that shortfall reaches $1$.

\paragraph{Loss.} The axiom is that the true order scores at least $m = 0.1$ above every other order: for each of the $119$ orders $\sigma \neq \mathrm{id}$ we assert $\bigl(\text{Score}(\sigma) \vee m\bigr) \to \text{Score}(\mathrm{id})$, with $\vee$ of Eq.~\ref{eq:luk}. Its cost (Eq.~\ref{eq:contra}) is $\relu\bigl(\text{Score}(\sigma) + m - \text{Score}(\mathrm{id})\bigr)$ whenever $\text{Score}(\sigma) \le 1 - m$; averaged over $\sigma$,
\begin{equation}
\mathcal{L}(\theta) \;=\; \frac{1}{119} \sum_{\sigma \neq \mathrm{id}}
    \relu\bigl(\text{Score}(\sigma) + m - \text{Score}(\mathrm{id})\bigr).
\label{eq:order_loss}
\end{equation}
AdamW, $50$ epochs, $5$ seeds. Each run's checkpoint is chosen on in-domain validation, so the in-domain numbers below are slightly optimistic and the Story Cloze numbers a held-out estimate.

\paragraph{Variants and metrics.} The minimal model consists of Eqs.~\ref{eq:order_C}--\ref{eq:order_loss}. Two variants remove one part each: no margin ($m = 0$), and a symmetric relation ($P_{\!\text{src}} = P_{\!\text{tgt}}$, $12{,}290$ parameters). One adds a part: \emph{+multi-formula} adds the formulas ``always coherent'' ($G$) and ``always eventually the end'' ($GF$) next to Until. The position classifier ($1{,}925$ parameters) predicts each sentence's position from its embedding alone. We count the fraction of whole stories put in the right order and the fraction of sentence pairs in the right relative order; an order that misplaces one sentence still gets most pairs right. Chance is $1/120$ for stories and $0.5$ for pairs. Three criteria were declared before running (code package): the minimal model reaches pairwise accuracy $\geq 0.78$ in domain and $\geq 0.74$ on Story Cloze, the model without margin stays below $0.78$, and the symmetric one below $0.70$, on every seed.

\begin{figure}[t]
\centering
\includegraphics[width=\linewidth]{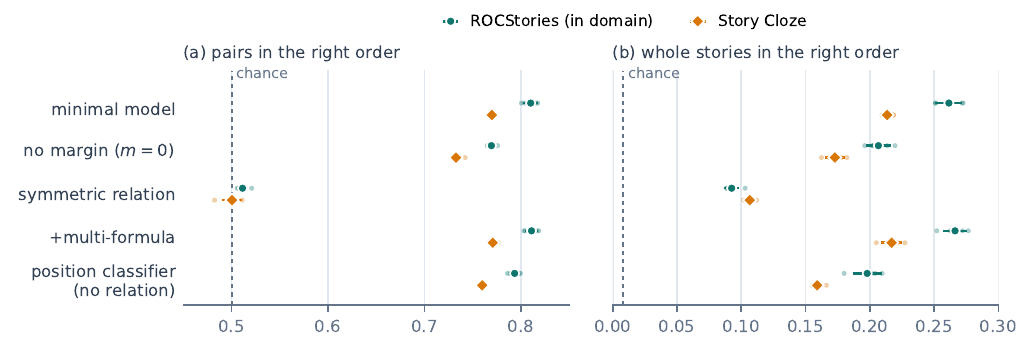}
\caption{Ablation over five seeds. (a) Fraction of sentence pairs and (b) fraction of whole stories put in the right order, in domain and on Story Cloze. Small dots: seeds; markers and bars: mean $\pm$ sample s.d. The position classifier has no relation between sentences.}
\label{fig:order_ablation}\label{tab:order_ablation}
\end{figure}

\paragraph{Results.} All three criteria hold on every seed (Fig.~\ref{fig:order_ablation}). Without the margin, exact recovery drops from $0.261$ to $0.207$ in domain. A symmetric relation hurts most: it gives an ordering and its reverse the same score, so the pairwise score falls to chance; the remaining exact score comes from orderings that are right up to reversal. \textsf{+multi-formula} changes nothing beyond seed noise. Finally, a classifier that predicts each sentence's position from its embedding alone, with no relation between sentences, comes within $0.02$ of the minimal model on pairs but recovers fewer whole stories ($0.198$ against $0.261$ in domain). Where a sentence typically sits explains most of the pairwise accuracy; the relation between sentences mainly adds getting the whole chain right.

\paragraph{Interpreting the relation.} Each prediction is computed from the $25$ entries of $A_\theta$ alone (Fig.~\ref{fig:order_interpretability}). On a misordered example one can see which edge the model over-trusted (solid border) and which it under-trusted (dashed box); the margin $\Delta$ is the amount by which the model ranked a wrong order above the truth. Any other LTL property of the recovered chain can be evaluated on the same entries without retraining.

\vspace{-1em}\section{Deontic: Traffic Norms}\vspace{-0.25em}
\label{sec:deontic}\suppressfloats[t]

Deontic logic reasons about what is obligatory and what is permitted, and its relation says which next situations an agent may move to: $\Box\phi$ (``$\phi$ is obligatory'') holds if $\phi$ is true after every permitted move, and $\Diamond\phi$ (``$\phi$ is permitted'') if it is true after some permitted move. We observe compliant drivers at a pedestrian crossing with a traffic light and have one rule written by a domain expert; from these we want a model of the hidden norm that can judge drivers who may not obey (Fig.~\ref{fig:traffic_scenario}). The case study shows two uses of axioms: a written rule that reaches situations the data never contain, and seriality (axiom D), without which obligations become vacuous.

\begin{figure}[!h]
\centering
\begin{tikzpicture}[font=\small, >=Stealth, x=1cm, y=1cm]
\fill[slate50] (0,0) rectangle (10.6,1.5);
\draw[slate700, line width=0.6pt] (0,0) -- (10.6,0) (0,1.5) -- (10.6,1.5);
\draw[slate700, line width=1.6pt] (2.55,0) -- (2.55,1.5);
\foreach \y in {0.12,0.42,0.72,1.02} \fill[white, draw=slate500, line width=0.3pt] (3.0,\y) rectangle ++(1.3,0.2);
\draw[slate500, dashed] (4.9,0) -- (4.9,1.5) (7.5,0) -- (7.5,1.5);
\node[font=\scriptsize, text=slate500] at (1.4,-0.28) {stop line S};
\node[font=\scriptsize, text=slate500] at (3.65,-0.28) {crosswalk C};
\node[font=\scriptsize, text=slate500] at (6.2,-0.28) {intersection box B};
\node[font=\scriptsize, text=slate500] at (9.0,-0.28) {exit E};
\node[draw=teal700, fill=teal50, rounded corners=2pt, minimum width=9mm, minimum height=5mm, font=\scriptsize] (car) at (1.5,0.75) {car};
\draw[->, teal700, line width=0.9pt] (car.east) -- node[above, font=\scriptsize] {advance} (2.95,0.75);
\draw[->, teal700, line width=0.9pt] (car.north west) to[out=150, in=-150, looseness=5] node[left, font=\scriptsize] {stay} (car.south west);
\node[draw=slate700, fill=slate200, minimum width=3.2mm, minimum height=7mm, inner sep=0pt] (tl) at (2.55,2.35) {};
\fill[rose700] (2.55,2.52) circle (1.1mm);
\fill[white] (2.55,2.18) circle (1.1mm);
\node[font=\scriptsize, text=slate500, anchor=east, align=right] at (2.25,2.35) {light:\\ red or green};
\node[font=\scriptsize, anchor=south west, align=left] (ped) at (3.95,1.95) {nearest pedestrian at distance $d \in \{0,\dots,5\}$\,m from the crosswalk\\ ($0$: on the crosswalk, $5$: nobody)};
\draw[->, slate700, line width=0.7pt] (3.85,2.0) -- node[pos=0.45, left, font=\scriptsize] {$d$} (3.65,1.62);
\node[font=\scriptsize, text=slate500, anchor=south west, align=left] at (3.95,2.75) {emergency vehicle approaching: yes / no\\ vehicle type: car, van, truck (bus held out)};
\end{tikzpicture}
\caption{The traffic scenario. A world is the situation of one car. Compliant drivers are observed (they enter the crosswalk only on green, with no pedestrian closer than $3$\,m and no emergency vehicle, and never stop on it); the norm itself is hidden. The model learns which of the two possible moves is permitted in every situation, including the roughly $81$ situations that never occur in the demonstrations, from the demonstrations plus one written rule.}
\label{fig:traffic_scenario}
\end{figure}
\begin{glance}{Setup}
Worlds & The situation of one car ($|W| = 384$), described by \\
 & position: stop line S, crosswalk C, intersection box B, exit E; \\
 & light: green or red; emergency vehicle: yes or no; \\
 & distance $d\in\{0,\dots,5\}$\,m of the nearest pedestrian ($0$: on the crosswalk, $5$: none); \\
 & vehicle type: car, van, truck (bus held out). \\
Moves & A car stays or advances one position: successors stay$(s)$ and adv$(s)$. \\
 & What \emph{can} happen is known; what \emph{may} happen is learned. \\
Relation & $A_\theta[s,s']$ = $s'$ is a permitted successor of $s$. \\
 & Metric form $\sigma(f_\theta(x_s)^\top g_\theta(x_{s'}) + b)$, two encoders, 8 features per world. \\
Axioms & (i) Demonstrated moves are permitted; other moves in visited situations are not. \\
 & (ii) The written rule ``never be on the crosswalk with a pedestrian on it'': \\
 & \phantom{(ii)} $\Box\neg\textit{conflict}$, with $\textit{conflict}(s') = [\text{pos}(s')=\text{C} \wedge d(s')=0]$. \\
 & (D) Seriality $\Diamond\top$: some move is always permitted. \\
Loss & Binary cross-entropy for (i); $1 - U\bigl(\Box\neg\textit{conflict}(s)\bigr)$ for (ii); $1 - \Diamond\top(s)$ for (D). \\
 & Averaged over all $216$ worlds with a move and a training vehicle type, visited or not. \\
Output & Obligation $\Box\phi$ and permission $\Diamond\phi$ at any situation. \\
 & Violation score of a driver: $1 - \min_t A_\theta[s_t, s_{t+1}]$ over its moves. \\
\end{glance}

\paragraph{Data.} The hidden norm: at the stop line, waiting is always permitted, and entering the crosswalk only on green, with no pedestrian closer than $3$\,m and no emergency vehicle; a car must not stop on the crosswalk. We simulate $300$ compliant drivers per seed (about $1{,}570$ steps), in which pedestrians cross only on red and drivers never hesitate at green. About $81$ of the $216$ worlds with a move never occur, among them every green situation with a pedestrian closer than $3$\,m. The written rule (ii) is correct but incomplete: it covers a pedestrian on the crosswalk ($d=0$), not one at $1$--$2$\,m. The test has $2{,}000$ new drivers per seed, half compliant and half violating the norm, and pedestrians now also stand close to the crosswalk on green. A move counts as flagged when $A_\theta < 0.5$; seeds are $42$--$46$.

\paragraph{Models.} The MLNN learns $A_\theta$ from the demonstrations (i), the rule (ii) and seriality (D), and applies the rule as a mask, $A = A_\theta \odot (1 - \mathit{conflict}(s'))$, so that D can only be satisfied by a permitted move. (Adding D to the loss alone permits the forbidden move again in three of five seeds, because the soft $\Diamond$ sends almost all of its gradient to the move that is already most permitted.) We compare it with the same network trained on the demonstrations alone, a positive--unlabelled (PU) classifier since only permitted moves are observed; with that classifier when the rule is applied afterwards as a filter that sets $A = 0$ on conflict moves; and with a lookup that permits exactly the observed moves. The code package reports eight further variants.

\begin{figure}[t]
\centering
\includegraphics[width=\linewidth]{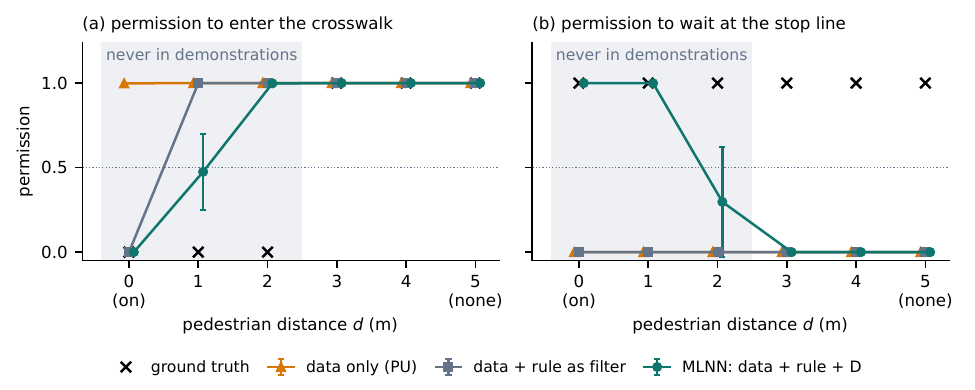}
\caption{Learned permission at the stop line, green light, no emergency vehicle, as a function of the pedestrian distance $d$ (mean $\pm$ s.d., five seeds). Crosses: ground truth. Shaded: distances that never occur at green in the demonstrations. For the MLNN the plotted relation is the masked one.}
\label{fig:deontic_profile}
\end{figure}

\begin{table}[t]
\centering
\caption{Test results (mean over five seeds; s.d.\ where non-zero). Flagged: share of moves with $A_\theta < 0.5$; violations with a pedestrian on the crosswalk or at $1$--$2$\,m. Dead ends: worlds, of $216$, with no permitted move. Drivers: AUROC between violating and compliant drivers, and the share of compliant drivers flagged.}
\label{tab:deontic_main}
\small
\setlength{\tabcolsep}{3.5pt}
\begin{tabular}{@{}lcccccc@{}}
\toprule
 & \multicolumn{2}{c}{Violations flagged (\%)} & Compliant waits & Dead & \multicolumn{2}{c}{Drivers} \\
\cmidrule(lr){2-3}\cmidrule(lr){6-7}
Model & on crosswalk & at $1$--$2$\,m & flagged (\%) & ends & AUROC & false alarms \\
\midrule
Lookup & $100$ & $100$ & $100$ & $81 \pm 3$ & $0.61$ & $0.79$ \\
Data only (PU) & $0$ & $0$ & $100$ & $0$ & $0.90 \pm 0.02$ & $0.38 \pm 0.01$ \\
Data + rule as filter & $100$ & $0$ & $100$ & $3$ & $0.93 \pm 0.01$ & $0.38 \pm 0.01$ \\
MLNN (data + rule + D) & $100$ & $29 \pm 21$ & $24 \pm 14$ & $0$ & $0.95 \pm 0.01$ & $0.20 \pm 0.05$ \\
\bottomrule
\end{tabular}
\end{table}

\paragraph{Results.} Near the data every model flags the violations, the lookup included. Away from the data they differ (Table~\ref{tab:deontic_main}, Fig.~\ref{fig:deontic_profile}). Trained on the demonstrations alone, the classifier never flags a car entering the crosswalk on green with a pedestrian on it: that situation never occurs, and the green light dominates. The lookup flags it only because it forbids every move it has not seen, compliant ones included. The rule fixes this, as a filter or in the MLNN, and in the MLNN it also reaches nearby situations: a pedestrian at $1$\,m lowers the permission to enter to about $0.5$ (Fig.~\ref{fig:deontic_profile}a), and $29\%$ of the $1$--$2$\,m violations are flagged. Seriality adds what a filter cannot. Where the rule forbids entering, waiting is the only move left, so the MLNN permits it without a single example (Fig.~\ref{fig:deontic_profile}b), while the filter leaves the car with no permitted move and flags every compliant wait there. The MLNN separates violating from compliant drivers best. One permission no model learns: waiting at green farther from a pedestrian is allowed by the norm, but no driver does it, and nothing in the rule or D implies it.

\paragraph{Frame axioms.} D is the frame axiom that matters for a permission relation, and it has to be stated during training. The others are not expected to hold for traffic moves (stopping on the crosswalk is forbidden, cars do not reverse), and B, 4 and 5 score at or below their shuffled null for every relation, the ground truth included.

\paragraph{Summary.} A written rule stated as a modal formula reaches situations the data never contain, and seriality lets the model infer a permission no driver demonstrated. Near the data, a lookup of the demonstrated moves does as well as any model.

\vspace{-1em}\section{Doxastic: Theory of Mind from Stories}\vspace{-0.25em}
\label{sec:tom}\label{sec:doxastic}\suppressfloats[t]

In Sect.~\ref{sec:knowledge} the relation was a kernel over hand-made features; here a neural network reads text and produces the relation. The task is theory of mind \citep{le2019revisiting, wu2023hitom}: people move through rooms and move objects, and the question is where one person thinks another person thinks an object is, nested up to four levels. The modality is belief, not knowledge: a person who left before the ball was moved believes something false.

\paragraph{From sentences to beliefs.} Take the story ``Anne and Bob entered the kitchen. The ball is in the box. Bob exited the kitchen. Anne moved the ball to the basket.'' Its sentences about the ball are two events, $e_1$ (the ball is placed in the box) and $e_2$ (it is moved to the basket). A world is a set of events that may have happened, and $\mathit{loc}_c$ holds in a world if its last event left the ball in container $c$. What a person believes depends on which events they saw. Bob left before $e_2$, so at the actual world $\{e_1, e_2\}$ he considers only $\{e_1\}$ possible, and $\Box_{\text{Bob}}\,\mathit{loc}_{\text{box}}$ holds: Bob thinks the ball is in the box. Nested beliefs apply $\Box$ again: ``Anne thinks Bob thinks the ball is in the box'' is $\Box_{\text{Anne}}\Box_{\text{Bob}}\,\mathit{loc}_{\text{box}}$, true because Anne saw both events and Bob only the first.

\begin{figure}[tp]
\centering
\resizebox{\textwidth}{!}{%
\begin{tikzpicture}[font=\sffamily\small, >=Stealth,
  src/.style={rectangle, rounded corners=1.5mm, draw=slate700, fill=slate50, line width=0.6pt, align=left, inner sep=4pt, text width=5.2cm},
  fixed/.style={rectangle, rounded corners=1.5mm, draw=slate500, fill=white, line width=0.6pt, align=center, inner sep=3pt, text width=5.3cm},
  learned/.style={rectangle, rounded corners=1.5mm, draw=teal700, fill=teal50, line width=0.9pt, align=center, inner sep=3pt, text width=5.3cm},
  world/.style={circle, draw=teal700, fill=teal50, line width=0.8pt, minimum size=1.05cm, inner sep=1pt, align=center, font=\sffamily\scriptsize},
  flow/.style={->, line width=0.8pt, draw=slate700},
  lab/.style={font=\sffamily\scriptsize\itshape, text=slate700},
  rel/.style={->, line width=1.1pt, draw=rose700},
  grad/.style={->, line width=1.2pt, dashed, draw=amber600}]
\node[src] (story) at (0,1.15) {\textbf{Story:} Anne and Bob entered the kitchen. The ball is in the box. Bob exited the kitchen. Anne moved the ball to the basket.};
\node[src] (q) at (0,-1.0) {\textbf{Question:} Where does Anne think Bob thinks the ball is?};
\node[fixed] (ev) at (7.2,1.75) {\textbf{events:} $e_1$ ball in box, $e_2$ ball to basket};
\node[learned] (obs) at (7.2,0.55) {\textbf{who saw what:} Anne $e_1, e_2$;\ Bob $e_1$};
\node[fixed] (fm) at (7.2,-1.0) {\textbf{formula:} $\Box_{\text{Anne}}\,\Box_{\text{Bob}}\,\mathit{loc}_c$};
\draw[flow] (story.east) ++(0,0.3) -- ++(0.35,0) |- node[lab, pos=0.75, above] {pattern matching} (ev.west);
\draw[flow] (story.east) ++(0,-0.3) -- ++(0.35,0) |- node[lab, pos=0.75, below] {network} (obs.west);
\draw[flow] (q.east) -- node[lab, above] {pattern matching} (fm.west);
\node[world] (w2) at (11.9,1.15) {$\{e_1, e_2\}$\\ basket};
\node[world] (w1) at (14.1,1.15) {$\{e_1\}$\\ box};
\draw[rel] (w2) edge[out=135, in=45, looseness=5] node[left=2pt, font=\sffamily\scriptsize, text=rose700] {Anne} (w2);
\draw[rel] (w2) -- node[below, font=\sffamily\scriptsize, text=rose700] {Bob} (w1);
\draw[rel] (w1) edge[out=135, in=45, looseness=5] node[right=2pt, font=\sffamily\scriptsize, text=rose700] {Bob} (w1);
\node[font=\sffamily\scriptsize\bfseries, text=slate700] at (13.0,2.95) {Kripke model};
\draw[flow] (ev.east) -- (w2.north west);
\draw[flow] (obs.east) -- (w2.south west);
\node[fixed, text width=3.4cm] (ans) at (17.2,-1.0) {\textbf{answer: box}\\ Anne $\to \{e_1,e_2\}$, Bob $\to \{e_1\}$};
\draw[flow] (fm.east) -- node[lab, below] {evaluate on the Kripke model} (ans.west);
\draw[flow] (w1.east) -| (ans.north);
\draw[grad] (15.4,-0.25) -- node[lab, above, text=amber600, pos=0.45] {training: order-1 answers, gradient through soft $\Box$} (10.4,-0.25) -| (obs.south east);
\end{tikzpicture}}

\vspace{6pt}
\includegraphics[width=\linewidth]{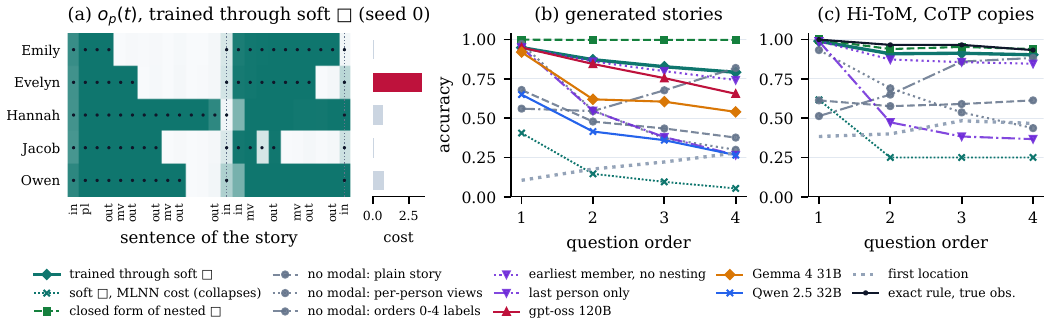}
\caption{Theory of mind from stories. Top: from story to answer; only who saw what is learned. Bottom: (a) the learned $o_p(e)$ for each sentence of a test story (teal high; dots: where each person truly was); right, each person's largest contradiction cost when one of Evelyn's reports is replaced by a wrong container (crimson). (b) Accuracy by question order on generated stories; neural models and the first-location rule are means over $5$ seeds of $600$ test stories, LLMs are on $200$ stories. (c) The same on Hi-ToM's $60$ chain-of-thought (CoTP) copies per order, with the exact rule on the true observations. ``MLNN cost'': trained with the interval cost $1-U$ / $L$ instead of the midpoint.}
\label{fig:tom}
\end{figure}

The events are found in the story by matching its sentence templates; what is not given is who saw which event. A small network reads the story from each person's point of view. Every sentence, with that person's name replaced by ``you'', is embedded by a frozen sentence encoder, and a recurrent layer over the embeddings outputs, for each sentence, the probability $o_p(e)$ that person $p$ saw its event. These probabilities define $p$'s accessibility relation (the formula is given below). The network outputs who saw what rather than the relation itself; this keeps the relation a belief relation over all worlds, so that it composes when $\Box$ is nested. The network thus learns from the text which worlds are accessible, as the traffic example of Sects.~\ref{sec:foundations}, \ref{sec:method} and~\ref{sec:deontic} learns which next situations are permitted; there the asserted formula is a written rule, here it is the first-order answers. Differentiable modal logic connects the relation to the answers. In training, only first-order answers are given, as formulas such as $\Box_{\text{Bob}}\,\mathit{loc}_{\text{box}}$; the soft $\Box$ is differentiable in the relation, so a wrong answer sends its gradient back to who saw what, and no observation is ever labelled. At test time, a question of order $k$ is $k$ nested $\Box$s on the same learned relation, a composition the network never sees in training (Fig.~\ref{fig:tom}, top).

\begin{glance}{Setup}
Worlds & Events about object $o$: its placement, its moves, and each person's view \\
 & of the room on entering. A world $w$ is a subset of these events. \\
 & $\mathit{loc}_c(w)$: the last event in $w$ left $o$ in container $c$. \\
Relation & $A_p[w,w']$ = person $p$ considers $w'$ possible at $w$. \\
 & Built from an observer head $o_p(e) = P(p \text{ observed } e)$. \\
 & Rounded: $B_p(w) = w \cap W_p$, with $W_p$ the events $p$ observed. \\
Network & Each sentence rewritten from $p$'s point of view ($p \to$ ``you''); \\
 & a frozen encoder (MiniLM), a GRU with $64$ units, one logit per sentence. \\
Loss & Order-1 answers only, through the soft $\Box$; no observation labels: \\
 & $\Box_p\,\mathit{loc}_{\mathrm{ans}}$ true, $\Box_p\,\mathit{loc}_c$ false for $c \neq \mathrm{ans}$; \\
 & $p$ believes the object is somewhere. \\
Queries & ``$p_1$ thinks $\dots$ $p_k$ thinks $o$ is in $c$'' $= \Box_{p_1}\cdots\Box_{p_k}\,\mathit{loc}_c$, \\
 & at the actual world; exact after rounding. \\
Baselines & The same network answering directly (three variants); \\
 & the learned head without nesting; ``the first location''; LLMs. \\
\end{glance}

\paragraph{Data.} Hi-ToM \citep{wu2023hitom} has $120$ stories with questions of orders $0$ (``where is $o$ really?'') to $4$. We keep the $600$ questions without deception (no claims or lies) and test on the $60$ chain-of-thought copies per order, whose labels follow one rule: a person observes what happens in a room while they are in it and sees where the room's objects are on entering, and a chain of people believes the last event that every person in it observed. The other prompt format's labels disagree with it on some questions. Because Hi-ToM is small, we also generate stories with the same templates, vocabulary and exact labels, in which the first location is the answer in $10$--$28\%$ of questions. Each seed trains on $3{,}000$ generated stories and tests on $600$; results are means over $5$ seeds.

\paragraph{From the network to the relation.} Person $p$ considers a world $w'$ possible at $w$ if $w'$ keeps exactly the events of $w$ that $p$ observed. With soft observations this becomes
\[
A_p[w,w'] = \min_{e\in w}\bigl(o_p(e)\ \text{if } e\in w',\ 1-o_p(e)\ \text{otherwise}\bigr) \quad\text{for } w' \subseteq w, \text{ and } 0 \text{ otherwise}.
\]
Rounded, it says that $p$ believes exactly the events $p$ saw, $B_p(w) = w\cap W_p$: a belief relation (KD45), reflexive only where $p$ missed nothing, like the faulty robot of Sect.~\ref{sec:knowledge}. Each first-order answer is scored with binary cross-entropy on the midpoint of the $\Box$ interval (the reason is given below). Two precautions are needed, because, as in Sects.~\ref{sec:knowledge} and~\ref{sec:deontic}, a $\Box$ target alone is satisfied by an empty relation. The relation takes the minimum over events, not the product, which would make every entry tiny; in a pilot the head learned to observe nothing. And we add the axiom that $p$ believes the object is somewhere, $\Box_p(\mathit{loc}_{c_1}\vee\dots\vee\mathit{loc}_{c_m}) = 1$, since believing it is nowhere satisfies every ``not in $c$'' target at once (also seen in the pilot; neither ablation is in the code package). At test time the observations are rounded and the exact $\Box$ of \tm{} answers the nested questions; on the rounded head it agrees with the closed form below on all $705$ questions checked (every 7th of a fixed set of $982$, five seeds).

\paragraph{A reference: the closed form.} For this relation the nested $\Box$ has a closed form: the answer is where the last event observed by all of $p_1,\dots,p_k$ left the object. Training the same head on its probabilistic version, the probability that this event is $e$ when observations are independent, shows how well the head does when trained on the exact answer formula instead of through the soft operator.

\paragraph{Results.} Trained through $\Box$ on first-order answers only, the model answers $0.95$ of the generated first-order questions and $0.79$ at order 4, and $0.99$ and $0.90$ on Hi-ToM, where the exact rule on the true observations scores $1.00$ and $0.93$ (Fig.~\ref{fig:tom}). Accuracy falls with the order because a question of order $k$ depends on what $k$ people saw; trained on the exact answer formula instead, the same head reaches $1.00$ at every order on generated stories, so most of the gap comes from the soft operator. Without the composition, the same network answering directly falls from $0.68$ to $0.38$, and LLMs given the rule degrade in the same way (gpt-oss 120B from $0.95$ to $0.66$).

The same contradiction cost points at an inconsistent person. When one person's report in each test story is replaced by a wrong container, the person with the largest cost is that person in $0.97 \pm 0.01$ of the $200$ stories, against $0.20$ by chance (five seeds; Fig.~\ref{fig:tom}a, right). The inconsistency is injected; Hi-ToM's stories with lies are not used.

\paragraph{The interval cost does not train.} Costing the answers by $1-U$ and $L$, as elsewhere (Sect.~\ref{sec:architecture}), drives the head to ``everyone observes everything'' on all five seeds although the cost itself prefers the truth, which is why we use the midpoint of the bounds.

\paragraph{Summary.} The relation is the output of a network trained only through $\Box$, and nested $\Box$ applies what it learned from first-order answers to fourth-order questions, which neither the same network trained on the same first-order answers without the composition nor any LLM tried does as well. This does not show that learning beats a symbolic tracker: given who was in the room, the exact rule is perfect by construction \citep{sclar2023minding}, and on templated stories a parser would recover the observations too. What the network learns is who observed what, from the text: the relation that classical modal inference has to be given before it can answer anything. Lies and private messages, the other half of Hi-ToM, would need a second relation, whom a person believes, and are not attempted.

\vspace{-1em}\section{Orchestration: A Cascade of Code Reviewers}\vspace{-0.25em}
\label{sec:orchestration}

An orchestrator decides which agents' outputs a system acts on. In our example LLMs review code, and the orchestrator decides whether to merge a pull request (PR). The reviewers differ in model, cost and what they see, and each reviewer's report narrows down what may be wrong with the PR. The team merges when only ``correct'' remains possible, rejects when only bugs remain, and otherwise asks a more expensive reviewer. These are the three cases of $\Box$ and $\Diamond$ over the hypotheses that remain.

\begin{glance}{Setup}
Worlds & The hypotheses about one PR, $H = \{\mathit{correct}\} \cup \{$six bug types$\}$. \\
 & Reports $R_i$ of reviewer $i$: correct, a bug type, or ``buggy'' without a type; \\
 & pass or fail for the continuous-integration (CI) test run. \\
Relation & One per reviewer: $A_i[h, r]$ = when the truth is $h$, reviewer $i$ may report $r$. \\
 & A \textsf{LearnableAccessibility} on $H \to R_i$. \\
Axiom & The observed report is possible under the true hypothesis: \\
 & $\Diamond_i(\mathit{report} = r_{\mathrm{obs}})$ at $h_{\mathrm{true}}$. \\
Loss & Its contradiction loss $1 - U$ (Eq.~\ref{eq:contra}), every hypothesis weighted equally; \\
 & plus a cost $\lambda = 0.099$ per link, an entry $A_i[h,r] = 1$ (\textsf{SparsityLoss}). \\
Output & The rounded relations, and the team decision of Eq.~\ref{eq:merge_rule}. \\
 & Axiom T, $A_i[h,h] = 1$ (reviewer $i$ names bug $h$ when it is present), \\
 & shows each reviewer's blind spots; reports and hypotheses share their labels. \\
Baselines & ``Merge iff CI passes''; the best single reviewer; majority vote; \\
 & logistic regression on all reports; \\
 & a supervised Dawid--Skene~\citep{dawid1979maximum}, asked in the same order until \\
 & its posterior passes a threshold chosen on the training folds. \\
\end{glance}

\begin{figure}[t]
\centering
\includegraphics[width=\linewidth]{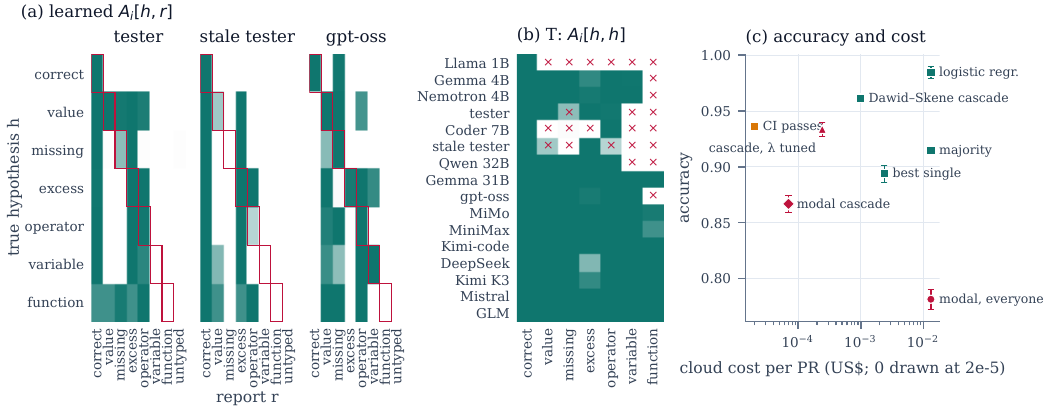}
\caption{\textbf{A modal cascade over seventeen code reviewers} (HumanEvalFix, $328$ PRs). (a) Learned relations $A_i[h,r]$ of three reviewers: the share of the $125$ fits ($25$ runs $\times$ $5$ folds) keeping each link (boxes: the diagonal $h = r$). (b) Axiom T, $A_i[h,h]$, for every LLM reviewer; $\times$ marks a blind spot. (c) Accuracy against cloud cost per PR (mean $\pm$ s.d.\ over $5$ fold splits $\times$ $5$ starts).}
\label{fig:code_review}
\end{figure}

\paragraph{Pull requests and reviewers.} HumanEvalFix~\citep{muennighoff2023octopack} has $164$ Python functions, each with its correct version and a version with a human-written bug of one of six types (value misuse, missing logic, excess logic, operator misuse, variable misuse, function misuse). This gives $328$ PRs, and hidden test suites confirm every label. Each reviewer sees one PR (the docstring and the diff) and answers a verdict, a bug type (or none) and a one-line reason, at temperature $0$. The cheapest reviewer is the CI run, which executes the examples in the function's docstring as tests; it is free, catches $143$ of the $164$ bugs and never fails a correct PR. Then come seven local models and nine Ollama Cloud models (all named in Fig.~\ref{fig:code_review}b), among them Qwen2.5 7B seeing the CI run on the PR (the \emph{tester}) or on the correct version (the \emph{stale tester}, whose evidence is out of date). The $2{,}952$ cloud reviews cost US\$4.30. The goal is a merge decision, learned from labelled PRs, that is right on unseen functions and asks few expensive reviewers.

\paragraph{Learning the relations.} The training data are labelled PRs: for each, the true hypothesis (correct, or which bug) and every reviewer's report. For each reviewer we learn which reports are possible under each hypothesis. The axiom asserts that the observed report is possible under the true one, and a cost per link keeps only the reports the reviewer gives on at least about $10\%$ of the PRs of a type. Nothing else is learned: the team decision below is a fixed rule. At test time the truth is unknown: the cascade sees only the reports and uses the learned relations on functions it was not trained on. An answer without a verdict, or a task without CI tests, rules nothing out. $\tau$ is annealed from $0.3$ to $0.01$, and the relation is rounded at $0.5$.

\paragraph{The team decision.} For reports $r_1, \dots, r_k$, the hypotheses still possible are $C(h) = \bigwedge_i A_i[h, r_i]$. We add a team world whose accessible worlds are exactly these hypotheses, and write $\Box_C$, $\Diamond_C$ for the operators at it. With the exact operators
\begin{equation}
\begin{gathered}
\text{reject iff } \Box_C\, \neg\mathit{correct}, \qquad \text{merge iff } \Box_C\, \mathit{correct},\\
\text{don't know iff } \Diamond_C\, \mathit{correct} \wedge \Diamond_C\, \neg\mathit{correct}.
\end{gathered}
\label{eq:merge_rule}
\end{equation}
If $C$ is empty, both boxes hold vacuously: the reports are inconsistent with every hypothesis. The cascade asks reviewers from cheap to expensive (CI, local models, cloud models by measured cost) only while the team does not know. A report that would empty $C$ is set aside and counted, and a PR still undecided after the last reviewer is rejected. Since $C$ only shrinks, a box that holds keeps holding: once the team knows, no later report changes its mind. So the cascade may stop early: asking every reviewer, and setting aside the same inconsistent reports, gives exactly the same decisions at $190$ times the cloud cost. Without the set-aside rule, intersecting all reports fails often: in half of the PRs some reviewer gives a report its relation calls impossible, and accuracy drops to $0.781$.

\paragraph{Results.} The learned relations show what each report is worth (Fig.~\ref{fig:code_review}a). When the CI run fails, the PR cannot be correct, so the team rejects it at once; this alone settles $43.6\%$ of the PRs. gpt-oss rarely approves a buggy PR, so its approval is possible only under \emph{correct}, and the team merges. Its most common false alarm, ``missing logic'', also occurs on correct PRs, so it settles nothing. The testers' approvals never settle anything, because they also approve buggy PRs of almost every type. The relations also show blind spots (Fig.~\ref{fig:code_review}b): no local model, and not gpt-oss, names function-misuse bugs.

On functions it has never seen, the cascade decides $0.867$ of the PRs correctly and spends almost nothing on cloud models (Fig.~\ref{fig:code_review}c), because most PRs are settled by the free CI run or a local model. Other methods are more accurate. Merging whenever CI passes reaches $0.936$, and logistic regression on all reports $0.985$, but it asks every reviewer. The cascade's mistakes have one cause. After rounding, a report that a cheap reviewer gives on $9\%$ of correct PRs counts as impossible for a correct PR, so a single such false alarm rules out \emph{correct} for good.

The modal decision brings three benefits: a ``don't know'' that tells the team when to pay for more evidence, a decision that never reverses, so the team can stop early, and a readable profile of each reviewer. It does not give probabilities. Where accuracy matters most, keep the relation graded, or use it only to choose whom to ask.

\vspace{-1em}\section{Coordination: Drones That Map and Check Each Other}\vspace{-0.25em}
\label{sec:communication}\suppressfloats[t]

Five drones fly over terrain with obstacles that nobody knows in advance, and must carry a baton from a start $S$ to a goal $G$ in a relay of three legs. Three drones are healthy. One is \emph{inaccurate}: it is honest, but its position, the drones it sees and its map are off by about $0.7$. One is an \emph{attacker}: it broadcasts a false position beside the obstacle on the course and reports the area around it as clear, also over the obstacle. At every step each drone broadcasts its position, the drones it sees, and the map cells it sees as free or blocked. From these messages the team must decide whom to believe, build a map, and plan the relay around the obstacles. Since every message is broadcast, every drone can run the same computation.

\begin{glance}{Setup}
Worlds & Positions $y$ in a $10\times10$ area and its map cells; three unknown obstacles. \\
 & Five drones, six steps; at each step a drone broadcasts position, sightings and map cells. \\
Propositions & ``a drone is at $y$''; ``cell $c$ is blocked''. \\
Relation & $V_i(x, y)$ = drone $i$ at $x$ would see $y$: $\sigma\bigl(c\,(r_i - |x - y|)\bigr)$, zero behind an obstacle. \\
Operators & Knowledge: $K_i\phi$ = $\Box$ over the positions drone $i$ sees. \\
 & Belief: $B_a K_b\,\phi$, what drone $b$ tells drone $a$. \\
Learned & The five ranges $r_i$; a trust value $t_b \in [0,1]$ per drone, refitted after each step. \\
Axioms & (i) ``I see $j$'' asserts $V_i(x_i, x_j)$, ``I do not see $j$'' its negation (calibration flights). \\
 & (ii) A belief does not contradict knowledge: $B_a K_b\,\phi \wedge K_a\neg\phi$ is a contradiction, \\
 & \phantom{(ii)} for $\phi$ = ``$b$ is at $p$'', ``a drone is at $q$'', ``cell $c$ is blocked'' (or free). \\
 & (iii) ``$a$ and $b$ are both right'', for every pair; distrusting a drone has a cost. \\
Loss & (i) $1 - V$ or $V$. (ii) The size $M_{ab}$ of the failed checks; errors of about one unit cost little. \\
 & (iii) $\sum_{a<b} M_{ab}\, t_a t_b + \lambda \sum_b (1-t_b)^2$, $\lambda = 7$. \\
Queries & Map: a cell is free or blocked by the trust-weighted vote; unknown ($\Diamond$ of both) if no trusted drone saw it. \\
 & Handover from $a$ to $b$ at $m$: $K_a K_b\,m$, so only trusted drones carry the baton. \\
Output & Trust after each step; the team's map; the route from $S$ to $G$ and its three carriers. \\
Baselines & Believing everyone; dropping the drone in most failed checks; \\
 & the three drones with the fewest failed checks. \\
\end{glance}

\paragraph{Learning what each drone sees.} On calibration flights, each drone states for every other drone whether it sees it. ``I see $j$'' asserts $V_i(x_i, x_j)$ true and ``I do not see $j$'' asserts it false, and the contradiction loss of Sect.~\ref{subsec:loss} learns the five ranges, which are off by $0.04$ on average.

\paragraph{Knowledge, belief, and the checks.} What drone $a$ sees is knowledge. In particular, $K_a(\text{no drone at } y)$ is $\Box$ over the positions $a$ can see: $V_a(x_a, y)$ holds and $a$ sees nothing near $y$. What another drone $b$ says is only belief, $B_a K_b$. A check fails when a belief contradicts knowledge: $b$ says it is at $p$ and $K_a(\text{no drone at } p)$; $b$ says it sees a drone at $q$ and $K_a(\text{no drone at } q)$; or $a$ and $b$ report the same map cell differently. Each failed check has a size, its contradiction loss. The checks tolerate errors of about one unit, so an inaccurate report costs a little and a false one a lot.

\paragraph{Trust.} After each step the team refits the trust values by gradient descent on loss (iii) of the table, where $M_{ab}$ is the size of the failed checks between $a$ and $b$ so far. Its first term is the contradiction loss of asserting ``$a$ and $b$ are both right'' (with the product as conjunction); its second is the cost of distrusting a drone. A drone that contradicts everyone loses its trust; a drone that contradicts a little keeps part of it.

\paragraph{The map and the relay.} The team's map is the trust-weighted vote of the reports. A cell that trusted drones saw is free or blocked by their vote; a cell that no trusted drone saw is unknown, both free and blocked remaining possible ($\Diamond$). The route from $S$ to $G$ is planned on this map: blocked cells and a margin around them are excluded, unknown cells cost more. The route is split into three legs. A handover from $a$ to $b$ needs $K_a K_b$ of the meeting point, $a$ must know that $b$ knows where to meet, so the three most trusted drones carry the baton.

\paragraph{Results.} Over $100$ rounds, the relay reaches $G$ in $0.99$ of the rounds with the learned trust, against $0.02$ when the team believes everyone and $1.00$ for a team that knows the roles and the terrain. Believing everyone hands the baton to the attacker, which claims to be on the course, or plans the route through the obstacle the attacker reported clear ($14$ rounds). Dropping the drone in most failed checks finds the attacker in $0.98$ of the rounds, but then trusts the inaccurate drone fully: it carries the baton, handovers miss, and the relay succeeds in $0.34$. The attacker's final trust averages $0.01$ and never exceeds $0.14$; the inaccurate drone keeps a median of $0.53$ (quartiles $0.37$--$0.66$), and the healthy drones never fall below $0.74$. A simple ranking does nearly as well: using the three drones with the fewest failed checks reaches $0.86$, but it keeps the inaccurate drone's map at full weight. The tolerance and $\lambda$ were set on this seed.

\begin{figure}[t]
\centering
\includegraphics[width=\linewidth]{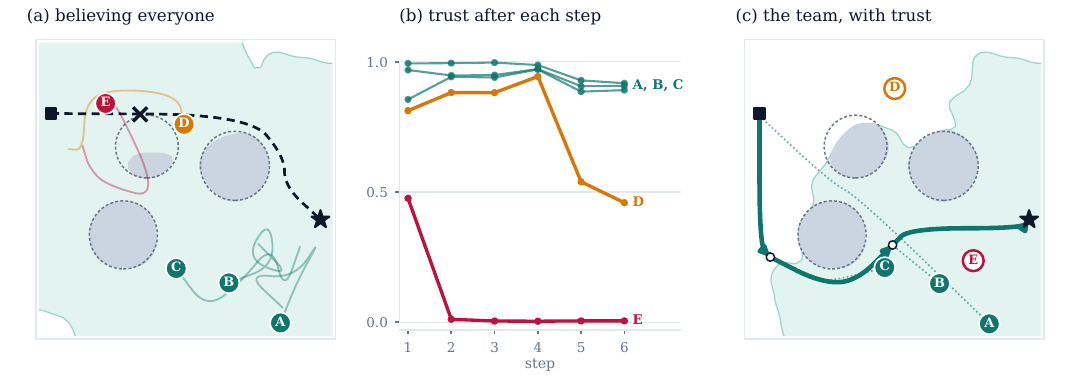}
\caption{One round (seed $42$). A, B and C are healthy, D is inaccurate, E is the attacker; the baton goes from $S$ (square) to $G$ (star). Dashed circles are the true obstacles; in the maps, grey is blocked, teal free, and white unknown. (a) Believing everyone. E broadcasts ``I am at $(2.3, 7.9)$, and around me it is all clear'', over the obstacle beside it, which almost vanishes from the shared map; the route crosses it ($\times$). (b) The trust learned after each step. A reports ``I would see E there, and I see nothing'', and B ``I see an obstacle where E says it is clear'', so E's trust falls below $0.1$ from the second step. D's reports are only slightly off and its trust settles at $0.46$; A, B and C stay near $0.9$. (c) With trust, E's map is dropped and D's counts half, the obstacle returns, and the route goes around it. A, C and B carry the baton (thick lines; dotted, their flights to the pickup points), each told where to meet by the previous carrier.}
\label{fig:relay}
\end{figure}

\paragraph{Summary.} The relation the team learns is epistemic: who would see what. A drone's own sensors give knowledge and the messages of others give belief; a contradiction between the two is a failed modal check, and its size grades the trust in the sender. The attacker loses all of it, the inaccurate drone about half, which a consistency check that only drops the worst drone cannot express (Fig.~\ref{fig:relay}). The method assumes a single attacker; several attackers that agree with each other could outvote the healthy drones.

\vspace{-1em}\section{Implementation}\vspace{-0.25em}
\label{app:from_scratch}\label{app:code}
\lstset{basicstyle=\scriptsize\ttfamily, aboveskip=3pt, belowskip=2pt}

The core of each study fits in a few lines. The listings use \tm{} 0.9 (\texttt{pip install torchmodal}) for the modal operators; Listings~\ref{lst:running}, \ref{lst:orchestrator} and~\ref{lst:trust} run as shown, and the others leave out loading the data, which is in the accompanying code package, one folder per section.

\subsection{The running example of Sect.~\ref{sec:method}}
\label{app:torchmodal_listing}
\begin{lstlisting}[caption={The running example of Sects.~\ref{subsec:soft_ops}--\ref{subsec:loss}.},label={lst:running}]
import torch
from torchmodal import functional as F

# Worlds: a waits at a red light, b drives off on green,
# c drives through on red.
red  = torch.tensor([1.0, 0.0, 1.0])
move = torch.tensor([0.0, 1.0, 1.0])
phi = F.implication(red, F.negation(move))    # red -> not move: [1, 1, 0]
pair = lambda x: torch.stack([x, x], dim=-1)  # interval [L, U] with L = U
A = torch.tensor([[1.0, 0.9, 0.6],            # row a: what may come next
                  [0.5, 1.0, 0.1],
                  [0.5, 0.9, 1.0]])
box = F.necessity(pair(phi), A, tau=0.1)      # [0.400, 0.402] at a
dia = F.possibility(pair(move), A, tau=0.1)   # [0.886, 0.905] at a

# The rule Box(red -> not move), asserted at a; train the logit of A[a,c].
logit = torch.logit(torch.tensor(0.6)).requires_grad_()
opt = torch.optim.Adam([logit], lr=0.05)
for step in range(300):
    row = torch.stack([A[0, 0], A[0, 1], torch.sigmoid(logit)])
    U = F.necessity(pair(phi), torch.stack([row, A[1], A[2]]), tau=0.1)[0, 1]
    loss = 1 - U                              # asserting it true costs 1 - U
    opt.zero_grad(); loss.backward(); opt.step()
print(torch.sigmoid(logit), U)                # A[a,c] -> 0.03, U -> 1.00
\end{lstlisting}

\subsection{Knowledge from reports (Sect.~\ref{sec:knowledge})}
\begin{lstlisting}[caption={Knowledge case study (Sect.~\ref{sec:knowledge}), metric form; the lines that build the data are omitted. \texttt{DIFF\_K[k,s,t]} is $1$ if worlds $s$ and $t$ differ on shelf $k$; \texttt{XT[k,s]} is $x_k(s)$; \texttt{T\_box}, \texttt{T\_dia} mark the asserted $K$ and $\Diamond$ literals of the reported worlds. Output: \texttt{[[1,2,3,4],[3,4,5],[5,6,7,8],[2,7]]}.},label={lst:knowledge}]
lits = torch.cat([XT, 1 - XT])                            # x_k and not x_k, (16, 70)
w = torch.nn.Parameter(torch.full((4, 8), 1.87))          # softplus(w) = 2: "sees every shelf"
beta = torch.nn.Parameter(torch.ones(4))
opt = torch.optim.Adam([w, beta], lr=0.1)
for t in range(600):
    tau = 0.1 * 0.2 ** (t / 599)                          # lower tau from 0.1 to 0.02
    d = torch.einsum("ik,kst->ist", torch.nn.functional.softplus(w), DIFF_K)
    A = torch.sigmoid(beta[:, None, None] - d)            # A_i[s,s'], (4, 70, 70)
    Ab, P = A[:, None].expand(4, 16, 70, 70), lits.expand(4, 16, 70)
    box = F.necessity(P, Ab, tau=tau)                     # K_i l  at every world
    dia = F.possibility(P, Ab, tau=tau)                   # <>_i l at every world
    loss = (F.contradiction(T_box, box) + F.contradiction(T_dia, dia)) \
           / (T_box.sum() + T_dia.sum())
    opt.zero_grad(); loss.backward(); opt.step()
sees = torch.nn.functional.softplus(w) > beta[:, None]    # robot i sees shelf k
print([[k + 1 for k in range(8) if sees[i, k]] for i in range(4)])
\end{lstlisting}

\subsection{Ordering the sentences of a story (Sect.~\ref{sec:temporal})}
\begin{lstlisting}[caption={Sentence ordering (Sect.~\ref{sec:temporal}): the relation of Eq.~\ref{eq:order_C}, the Until score of Eq.~\ref{eq:order_score} and the loss of Eq.~\ref{eq:order_loss} for one story. \texttt{h}: the five sentence embeddings in the true order; scrambling the sentences permutes the $120$ scores but leaves the loss unchanged.},label={lst:temporal}]
from itertools import permutations
ORDERS = torch.tensor(list(permutations(range(5))))   # the 120 orders; ORDERS[0] = true order
P_src = torch.nn.Linear(384, 32, bias=False)          # two linear encoders
P_tgt = torch.nn.Linear(384, 32, bias=False)
alpha, beta = torch.nn.Parameter(torch.ones(())), torch.nn.Parameter(torch.zeros(()))
psi = torch.tensor([0., 0., 0., 0., 1.])              # "this is the last sentence"

def score(A, s):                                      # Until at the first sentence
    phi = torch.cat([A[s[:-1], s[1:]], torch.ones(1)])  # step t of order s is coherent
    return F.until(phi, psi, torch.eye(5))[0]

def loss(h, m=0.1):                                   # h: (5, 384)
    A = torch.sigmoid(alpha * P_src(h) @ P_tgt(h).T + beta)
    S = torch.stack([score(A, s) for s in ORDERS])
    return torch.relu(S[1:] + m - S[0]).mean()        # the true order wins by m
\end{lstlisting}

\subsection{Permissions with a written rule and seriality (Sect.~\ref{sec:deontic})}
\begin{lstlisting}[caption={Deontic case study (Sect.~\ref{sec:deontic}), the variant with the rule built into the relation. \texttt{XS}: features of each source world; \texttt{XT}: features of its two possible successors (stay, advance); \texttt{CONF}: conflict of each successor; \texttt{TR}: worlds of the training vehicle types; \texttt{net} is the two-encoder metric relation of Sect.~\ref{subsec:accessibility}.},label={lst:deontic}]
def box_U(phi, A):        # upper end U of the interval of O phi = Box phi
    P = torch.stack([phi, phi], -1)                    # point values as intervals [L, U]
    return F.necessity(P, A.unsqueeze(1), tau=0.1)[:, 0, 1]
dia = lambda phi, A: F.possibility(phi, A.unsqueeze(1), tau=0.1).squeeze(-1)  # P phi
opt = torch.optim.Adam(net.parameters(), lr=5e-3)
bce = torch.nn.BCELoss()
for step in range(3000):
    A_theta = net(XS, XT)                         # (N, 2) learned permission
    A = A_theta * (1 - CONF)                      # rule built into the relation
    # (i) demonstrated moves permitted, untaken moves in visited situations not
    loss = bce(A_theta[pr, pm], torch.ones(len(pr))) \
         + bce(A_theta[ur, um], torch.zeros(len(ur)))
    # (ii) O(not conflict) = Box(not conflict) at every training world, costed by U
    loss = loss + (1 - box_U(1 - CONF[TR], A_theta[TR])).mean()
    # (D) Diamond(top): some move is permitted, on the relation used for reasoning
    loss = loss + (1 - dia(torch.ones_like(A[TR]), A[TR])).mean()
    opt.zero_grad(); loss.backward(); opt.step()
\end{lstlisting}

\subsection{Observer head and nested belief (Sect.~\ref{sec:tom})}
\begin{lstlisting}[caption={Observer head and nested belief (Sect.~\ref{sec:tom}), with the loss of the closed-form reference; the main model instead applies binary cross-entropy to the midpoint of the bounds from \texttt{F.necessity}. \texttt{view[p]}: the sentence embeddings of the story rewritten from $p$'s point of view; \texttt{at[e]}: the sentence at which event \texttt{e} is observed; \texttt{dest[e]}: where it leaves the object.},label={lst:tom}]
gru, out = torch.nn.GRU(384, 64, batch_first=True), torch.nn.Linear(64, 1)
def observe(p):                                    # o_p(t) for every sentence t
    return torch.sigmoid(out(gru(view[p][None])[0])).squeeze()

def belief(chain, events):                         # box_p1 ... box_pk loc at the actual world
    keep = torch.stack([observe(p)[at[events]] for p in chain]).prod(0)
    later = torch.cat([torch.cumprod((1 - keep).flip(0), 0).flip(0)[1:], torch.ones(1)])
    return keep * later                            # P(the chain's last common event is e)

for story, p, events, ans in order1_answers:       # no observation labels
    P = belief([p], events)
    loss = -torch.log(P[dest[events] == ans].sum())
    opt.zero_grad(); loss.backward(); opt.step()
\end{lstlisting}

\vspace{-1em}\subsection{The modal cascade for code review (Sect.~\ref{sec:orchestration})}\vspace{-0.25em}
\label{app:orchestrator}

\begin{lstlisting}[language=Python,
  label={lst:orchestrator},
  caption={One reviewer's relation $A_i$ (Sect.~\ref{sec:orchestration}) learned from the axiom ``the observed report is possible under the true hypothesis'', and the three-valued team decision. \texttt{h}, \texttt{r}: long tensors of true hypotheses (0 = correct) and observed reports.}]
import torch
from torchmodal import SparsityLoss, functional as F, nn as tmnn
from torchmodal.verify import round_relation

def fit(h, r, nH=7, nR=8, lam=0.099, steps=400, seed=0):
    torch.manual_seed(seed); W = nH + nR           # worlds: hypotheses, then reports
    acc = tmnn.LearnableAccessibility(W, init_bias=2.0, reflexive=False)  # all links start near 1
    with torch.no_grad(): acc.logits.add_(0.5 * torch.randn(W, W))        # per-start noise
    mask = torch.zeros(W, W); mask[:nH, nH:] = 1   # only h -> r links
    phi = torch.zeros(len(r), W, 2); phi[torch.arange(len(r)), nH + r] = 1
    w = 1 / torch.bincount(h, minlength=nH)[h] / nH  # class-balanced
    opt, sparse = torch.optim.Adam(acc.parameters(), lr=0.1), SparsityLoss(lam * nR, False)
    for t in range(steps):
        tau = 0.3 * (0.01 / 0.3) ** (t / (steps - 1))   # anneal
        A = acc() * mask
        dia = F.possibility(phi, A.expand(len(r), W, W), tau=tau)
        loss = (w * (1 - dia[torch.arange(len(r)), h, 1])).sum() + sparse(A[:nH, nH:])  # 1 - U
        opt.zero_grad(); loss.backward(); opt.step()
    return round_relation((acc() * mask)[:nH, nH:].detach()).bool()

def decide(C):                       # C: (PRs, nH) hypotheses still possible
    B, nH = C.shape
    A = torch.zeros(B, nH + 1, nH + 1); A[:, 0, 1:] = C.float()
    ok = torch.zeros(B, nH + 1, 2); ok[:, 1] = 1        # h = correct
    bug = torch.zeros(B, nH + 1, 2); bug[:, 2:] = 1      # h != correct
    box = lambda p: F.necessity(p, A, mode="exact")[:, 0, 0] >= 0.5
    dia = lambda p: F.possibility(p, A, mode="exact")[:, 0, 0] >= 0.5
    out = torch.full((B,), -1)                         # -1: don't know
    out[box(bug) & dia(bug)] = 0; out[box(ok) & dia(ok)] = 1
    out[box(ok) & box(bug)] = -2                       # C empty: inconsistent
    return out

torch.manual_seed(1); h = torch.randint(0, 7, (300,))  # toy reviewer: right 80% of the time
r = torch.where(torch.rand(300) < 0.8, h, torch.randint(0, 8, (300,)))
A = fit(h, r)
print(A.int()); print(decide(A[:, r[:5]].T), h[:5])   # one reviewer's evidence: C = A[:, r]
\end{lstlisting}

\subsection{Trust among drones (Sect.~\ref{sec:communication})}
\begin{lstlisting}[caption={Trust (Sect.~\ref{sec:communication}): loss (iii) of the setup table, fitted by gradient descent. \texttt{M[a, b]}: the size of the failed checks between drones $a$ and $b$ so far, computed from their messages (\texttt{relay\_map.py} in the code package). On the toy matrix shown, the healthy drones keep $0.91$, the slightly inaccurate drone $0.61$ and the liar $0.003$.},label={lst:trust}]
import torch

def fit_trust(M, lam=7.0, steps=300):    # M[a, b]: size of the failed checks between a and b
    u = torch.full((len(M),), 2.0, requires_grad=True)       # t = sigmoid(u)
    opt = torch.optim.Adam([u], lr=0.1)
    for _ in range(steps):
        t = torch.sigmoid(u)
        loss = 0.5 * (M * t[:, None] * t[None, :]).sum() + lam * ((1 - t) ** 2).sum()
        opt.zero_grad(); loss.backward(); opt.step()
    return torch.sigmoid(u).detach()

M = torch.zeros(5, 5)                    # A, B, C healthy; D off by a little; E lies
M[3, :3] = M[:3, 3] = 2.0
M[4, :4] = M[:4, 4] = 10.0
print(fit_trust(M))                      # [0.91, 0.91, 0.91, 0.61, 0.003]
\end{lstlisting}

\vspace{-1em}\section{Limitations}\vspace{-0.25em}
\label{sec:limitations}

\paragraph{The axioms are given.} Training fits the relation to the axioms; it does not choose them. A wrong or incomplete axiom gives a relation that satisfies the wrong constraint, and nothing in the loss signals this. Errors of the components that compute the propositions (an encoder, a simulator, LLM verdicts) also become errors in the relation.

\paragraph{Untested entries.} An entry that appears in no violated axiom keeps its initial value, like the rows of unreported worlds in Sect.~\ref{sec:knowledge}. Which of the many relations with minimal loss is returned depends on the initialization and the optimizer, so a learned matrix should be interpreted together with the evidence behind each entry.

\paragraph{Training and baselines.} Two axioms that can hold together may pull against each other (Sect.~\ref{sec:deontic}), a $\Box$ target alone is satisfied by the empty relation (Sects.~\ref{sec:knowledge}, \ref{sec:tom}), a right cost may not train (Sect.~\ref{sec:tom}), and an imposed frame axiom cannot be checked (Sect.~\ref{sec:knowledge}). A simpler method does as well, or nearly as well, on every task itself (Table~\ref{tab:studies}); the modal model also gives a relation that answers questions it was not trained on.

\paragraph{No guarantees.} A high soft $\Box\phi$ implies a high hard $\Box\phi$ on the learned relation, but that relation is one of many that fit the data.

\paragraph{The worlds must be listed.} Every operator ranges over a finite set of worlds fixed before it is evaluated, so the relation is a matrix, with $O(|W|^2)$ memory and time per operator even in the metric form. The metric form gives a new world a row from its features (Sect.~\ref{subsec:accessibility}), but the world must still be listed: a system that creates processes during a run needs its worlds enumerated again, and a continuous state space must be discretized or sampled. Nesting multiplies the cost and the bias of the soft operators.

\vspace{-1em}\section{Conclusion}\vspace{-0.25em}
\label{sec:conclusion}

Differentiable modal logic makes the accessibility relation a parameter, learned by lowering the contradiction of axioms evaluated with soft $\Box$ and $\Diamond$, without labelling the relation. Learned from statements, sentence orders, behaviour with one written rule, stories processed by a network, the reports of LLM reviewers, or what drones see, it answers questions it was not trained on, nested ones included, and its frame axioms show properties such as the difference between knowledge and belief. On the tasks themselves simpler methods often do as well, and we report them.

\bibliography{references}

\end{document}